\documentclass[final,3p,times,sort&compress]{elsarticle}
\usepackage{comment}
\usepackage{amssymb}
\usepackage{amsthm}
\usepackage{amsmath}
\usepackage{lineno}
\usepackage{algorithm} 
\usepackage{algorithmic} 
\usepackage{multirow} 
\usepackage{lineno}
\usepackage{graphicx}
\usepackage{caption}
\usepackage{booktabs}
\usepackage{enumerate}
\usepackage{picinpar}
\usepackage{amsfonts}
\usepackage{stfloats}
\usepackage{flushend}
\usepackage{colortbl}
\usepackage{soul}
\usepackage{multirow}
\usepackage{tabularx}
\usepackage{threeparttable}
\usepackage{array}
\usepackage{makecell}
\usepackage{pifont}
\usepackage{color}
\usepackage{alltt}
\usepackage{subfigure}
\usepackage{setspace}
\usepackage{url}
\usepackage[colorlinks,linkcolor=red]{hyperref}
\usepackage{appendix}
\usepackage{bbm}
\usepackage{pdfpages}
\usepackage{pdflscape}
\usepackage{enumerate}
\usepackage{siunitx}
\usepackage{breakurl}
\usepackage{epstopdf}
\usepackage{pbox}
\usepackage{color}
\usepackage[dvipsnames]{xcolor}
\usepackage[skins]{tcolorbox}
\definecolor{b}{rgb}{0.00,0.00,1.00}    % b -- \textgreater blue
\definecolor{r}{rgb}{1.00,0.00,0.00}    % r -- \textgreater red

\usepackage{algorithm}
\usepackage{algorithmic}
\usepackage{adjustbox}

\journal{XXXX}
\newcommand{\chatdate}[2]{
        \begin{center}
        {\scriptsize \textbf{#1} #2}
        \end{center}
        \nopagebreak
}
\newenvironment{chat} {
    \newcommand\chatpersonOne[1]{
		\begin{tcolorbox}
				[enhanced, % 启用 tcolorbox 高级功能（允许阴影、圆角、附件等）
				before skip=-8pt, % 盒子前的垂直间距
				after skip=-14pt, % 盒子后的垂直间距
				left=2.8em,right=0em, % left / right → 盒子内部文字到边框的距离
				title={\includegraphics[width=0.6cm]{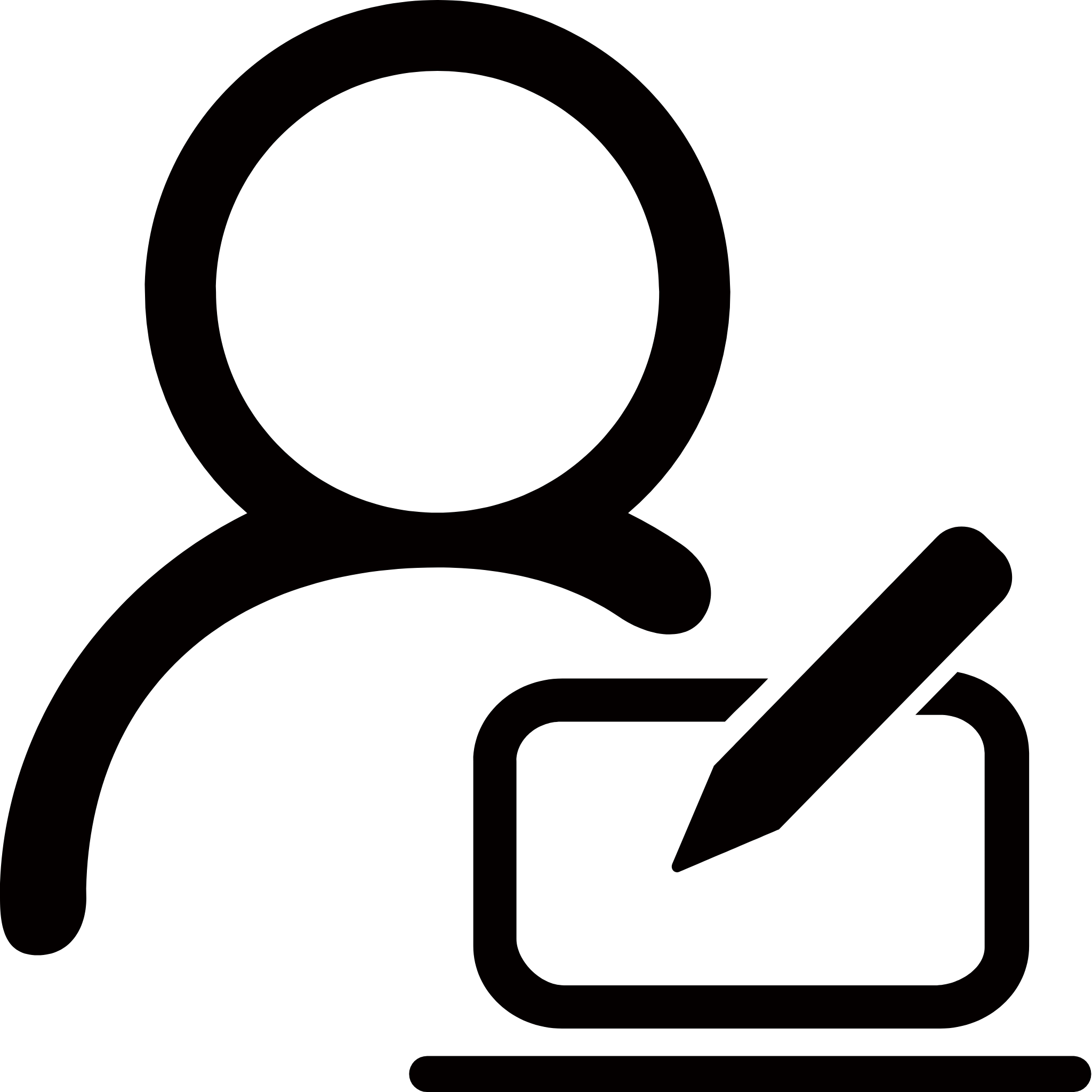}}, % title → 显示头像图片
				attach boxed title to top left={xshift=-2mm,yshift=-10mm}, % attach boxed title to top right → 头像在盒子右上角偏移位置
				opacitybacktitle=0.1, % opacitybacktitle → 头像背景透明度
				boxed title style={size=title,colback=white,colframe=white}, % boxed title style → 头像框的大小、颜色
				colframe=white, % colframe → 盒子边框颜色
				colback=Green!15!white, % colback → 盒子背景颜色
				halign=justify, % 两端对齐
				boxsep=1mm,  % 紧凑
				fontupper=\scriptsize\sffamily % fontupper → 盒子内文字字体大小、字体族
				]
		\nopagebreak
		##1
		\end{tcolorbox}
	}

    \newcommand\chatpersonTwo[1]{
        \begin{tcolorbox}
                [enhanced,
				before skip=-8pt, % 盒子前的垂直间距
				after skip=-14pt, % 盒子后的垂直间距
				left=0em,right=2.8em,
				title={\includegraphics[width=0.6cm]{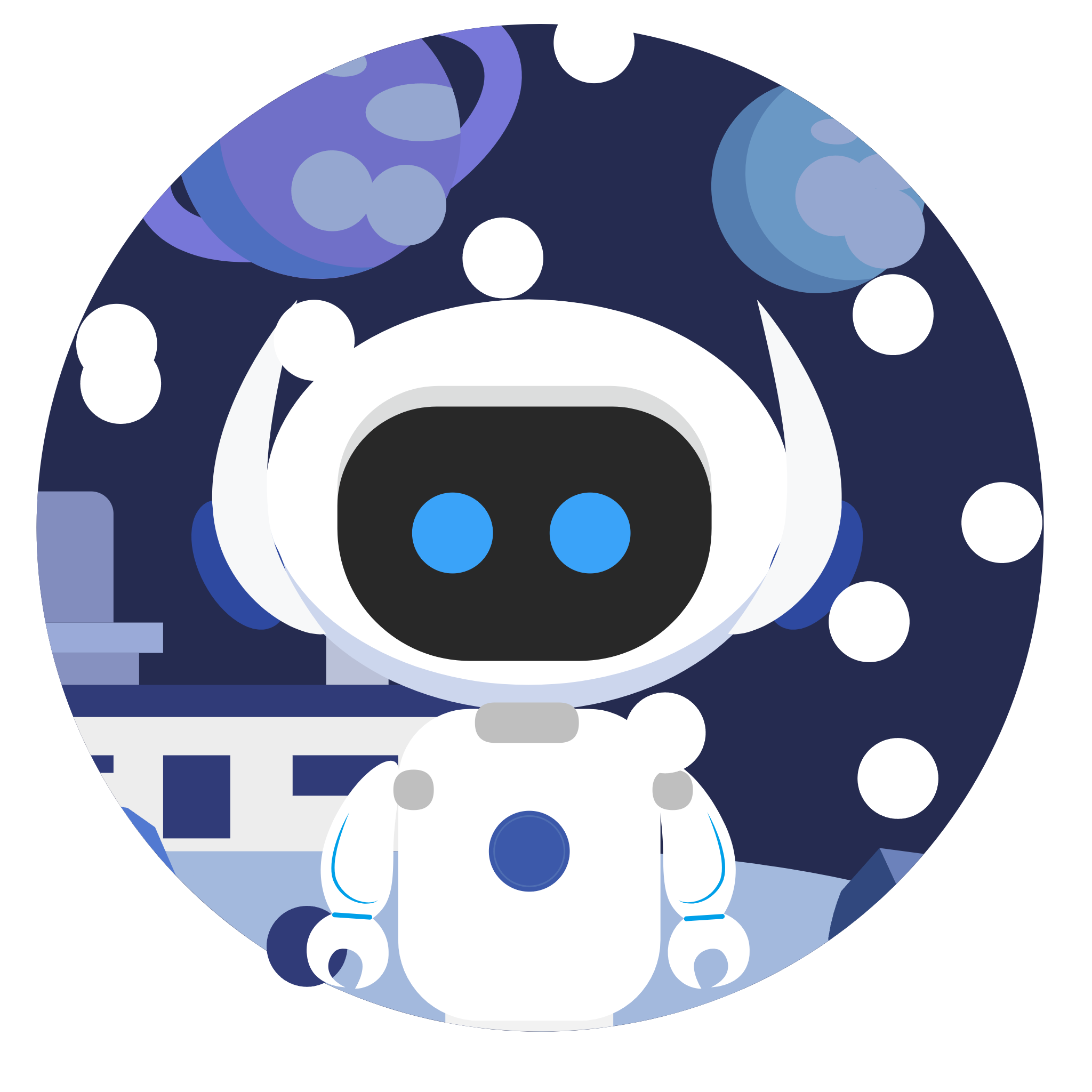}},
				attach boxed title to top right={xshift=2mm,yshift=-10mm},
				opacitybacktitle=0.1,
				boxed title style={size=title,colback=white,colframe=white},
				colframe=white,
				colback=blue!5!white,
				halign=justify, % 两端对齐
				boxsep=1mm,  % 紧凑
				fontupper=\scriptsize\sffamily
				]
          \nopagebreak
          ##1
        \end{tcolorbox}
    }

	\newcommand\chatpersonThree[1]{
        \begin{tcolorbox}
                [enhanced,
				before skip=-8pt, % 盒子前的垂直间距
				after skip=-14pt, % 盒子后的垂直间距
				left=0em,right=2.8em,
				title={\includegraphics[width=0.6cm]{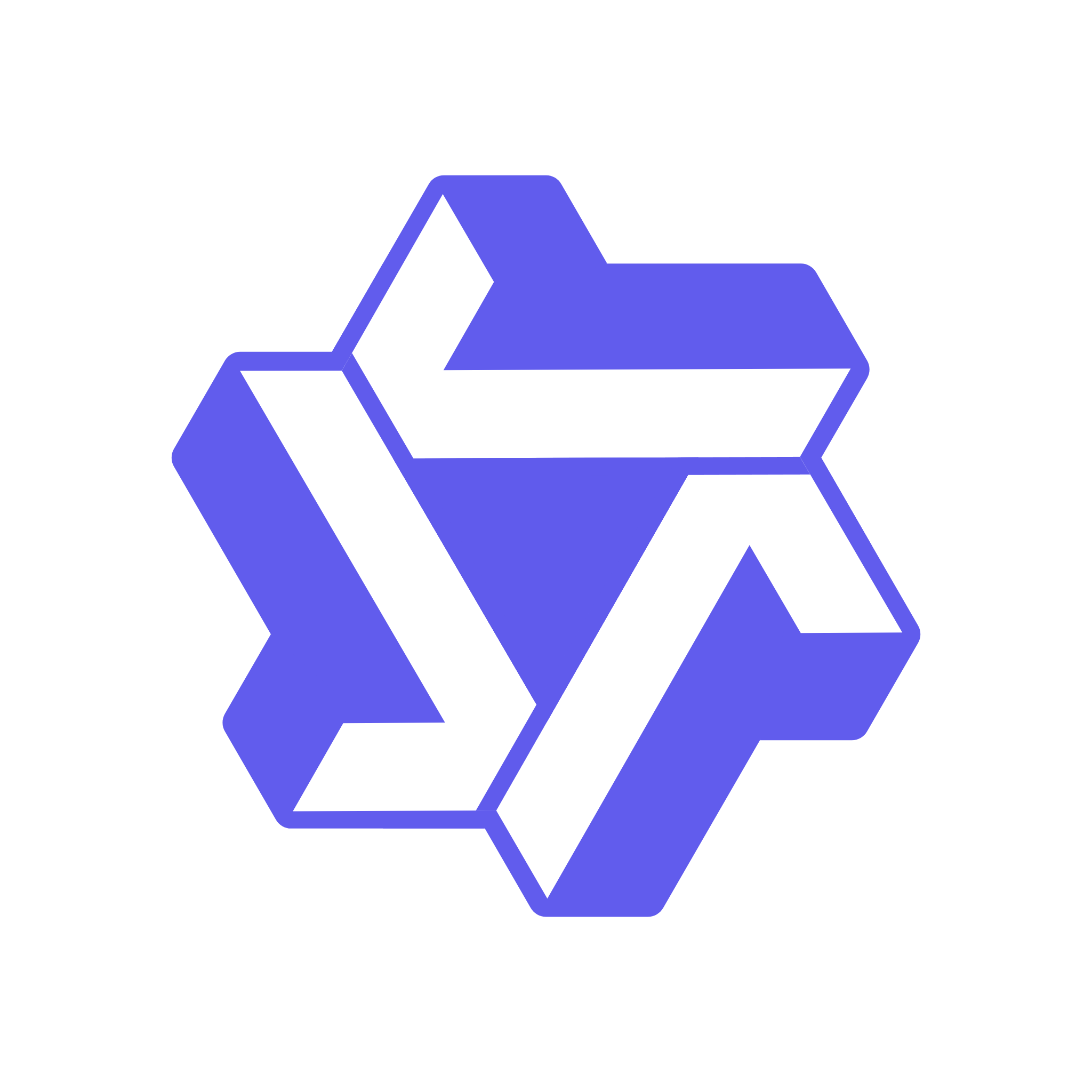}},
				attach boxed title to top right={xshift=2mm,yshift=-10mm},
				opacitybacktitle=0.1,
				boxed title style={size=title,colback=white,colframe=white},
				colframe=white!95!black,
				colback=white!95!black,
				halign=justify, % 两端对齐
				boxsep=1mm,  % 紧凑
				fontupper=\scriptsize\sffamily
				]
        \nopagebreak
        ##1
        \end{tcolorbox}
    }

}{
    \hangindent=0pt\hangafter=0
}

\begin{document}
	
%\list of figures
\captionsetup[figure]{labelfont={bf},labelformat={default},labelsep=period,name={Fig.}}	
		
\begin{frontmatter}

\title{Empowering All-in-Loop Health Management of Spacecraft Power System in the Mega-Constellation Era via Human-AI Collaboration
}

% \author[add1,add2]{Yi Di}%
% \ead{dy1999@stu.xjtu.edu.cn}%
% \author[add1,add2,1]{Zhibin Zhao\corref{cor1}}%
% \ead{zhaozhibin@xjtu.edu.cn}%
% \author[add3]{Fujin Wang}%
% \ead{wangfujin@kust.edu.cn}%
% \author[add4]{Xue Liu}%
% \ead{xueliu@stu.xjtu.edu.cn}%
% \author[add1,add2]{Jiafeng Tang}%
% \ead{tangjiafeng@stu.xjtu.edu.cn}%
% \author[add1,add2]{Jiaxin Ren}%
% \ead{r19981005@stu.xjtu.edu.cn}%
% \author[add1,add2]{Zhi Zhai\corref{cor1}}%
% \ead{zhaizhi@xjtu.edu.cn}%
% \author[add1,add2]{Xuefeng Chen}%
% \ead{chenxf@xjtu.edu.cn}%
% \ead{chenxf@xjtu.edu.cn}%
% \cortext[cor1]{Corresponding author: Zhibin Zhao and Zhi Zhai}%
% \address[add1]{National Key Lab of Aerospace Power System and Plasma Technology, Xi'an Jiaotong University, Xi'an 710049, P.R. China}%
% \address[add2]{School of Mechanical Engineering, Xi'an Jiaotong University, Xi'an 710049, P.R. China}%
% \address[add3]{Faculty of Mechanical and Electrical Engineering, Kunming University of Science and Technology, Kunming, 650550, P.R. China}%
% \address[add4]{Zhijian Laboratory, Rocket Force University of Engineering, Xi'an, 710025, P.R. China}%
% \footnotetext[1]{Lead Contact: Zhibin Zhao (zhaozhibin@xjtu.edu.cn). Present address: Xi'an Jiaotong University, Xi'an 710049, P.R. China}%

\author[]{Yi Di$^{1,2}$}%
\ead{dy1999@stu.xjtu.edu.cn}%
\author[]{Zhibin Zhao$^{1,2,*}$}%
\ead{zhaozhibin@xjtu.edu.cn}%
\author[]{Fujin Wang$^{3}$}%
\ead{wangfujin@kust.edu.cn}%
\author[]{Xue Liu$^{4}$}%
\ead{xueliu@stu.xjtu.edu.cn}%
\author[]{Jiafeng Tang$^{1,2}$}%
\ead{tangjiafeng@stu.xjtu.edu.cn}%
\author[]{Jiaxin Ren$^{1,2}$}%
\ead{r19981005@stu.xjtu.edu.cn}%
\author[]{Zhi Zhai$^{1,2,*}$}%
\ead{zhaizhi@xjtu.edu.cn}%
\author[]{Xuefeng Chen$^{1,2}$}%
\ead{chenxf@xjtu.edu.cn}%
\cortext[cor1]{Corresponding author: Zhibin Zhao (zhaozhibin@xjtu.edu.cn) and Zhi Zhai (zhaizhi@xjtu.edu.cn). Present address: Xi'an Jiaotong University, Xi'an 710049, P.R. China.}%
\footnotetext[1]{National Key Lab of Aerospace Power System and Plasma Technology, Xi'an Jiaotong University, Xi'an 710049, P.R. China}%
\footnotetext[2]{School of Mechanical Engineering, Xi'an Jiaotong University, Xi'an 710049, P.R. China}%
\footnotetext[3]{Faculty of Mechanical and Electrical Engineering, Kunming University of Science and Technology, Kunming, 650550, P.R. China}%
\footnotetext[4]{Zhijian Laboratory, Rocket Force University of Engineering, Xi'an, 710025, P.R. China}%

%%%%%%%%%%%%%%%%%%%%%%%%%%%%%%%%%%%%%%%%%%%%%%%%%%%%%%%%%%%%%%%%%%%%%%%%%%%%%%%%%%%%%%%%%%%%%%%%%%%%
%%%%%%%%%%%%%%%%%%%%%%%%%%%%%%%%%%%%%%%%%%% abstract %%%%%%%%%%%%%%%%%%%%%%%%%%%%%%%%%%%%%%%%%%%%%%%
\begin{abstract}
	It is foreseeable that the number of spacecraft will increase exponentially, ushering in an era dominated by satellite mega-constellations (SMC).
	The unprecedented scale and complexity introduce challenges to the health management of spacecraft, particularly spacecraft power systems (SPS), given their role in power supply and high failure rates.
	Providing health management for dozens of SPS and for thousands of SPS represents two fundamentally different paradigms.
	This work proposes a principle of aligning underlying capabilities (AUC principle) and develops SpaceHMchat, an open-source Human-AI collaboration (HAIC) framework for all-in-loop health management (AIL HM).
	SpaceHMchat serves across the entire loop of work condition recognition, anomaly detection, fault localization, and maintenance decision making, achieving goals such as conversational task completion, adaptive human-in-the-loop learning, personnel structure optimization, knowledge sharing, efficiency enhancement, as well as transparent reasoning and improved interpretability.
	Meanwhile, to validate this exploration, a hardware-realistic fault injection experimental platform is established, and its simulation model is built and open-sourced, both fully replicating the real SPS.
	The corresponding publicly available XJTU-SPS dataset includes four sub-datasets, involves 4 types of sub-tasks, 17 types of faults, over 700,000 timestamps.
	The experimental results demonstrate that SpaceHMchat achieves excellent performance across 23 metrics, such as 100\% conclusion accuracy in logical reasoning of work condition recognition, over 99\% success rate in anomaly detection tool invocation, over 90\% precision in fault localization, and knowledge base search time under 3 minutes in maintenance decision-making.

	% %%%%% The journal requires the abstract to be less than 150 words:
	% The satellite mega-constellations (SMC) era necessitates a focus on energy in space: spacecraft power systems (SPS).
	% Providing health management for dozens of SPS and for thousands of SPS represents two fundamentally different paradigms.
	% This work proposes a principle of aligning underlying capabilities (AUC principle) and develops SpaceHMchat, an open-source Human-AI collaboration (HAIC) framework for all-in-loop health management (AIL HM).
	% A hardware-realistic fault injection experimental platform is established to validate this exploration, and its simulation model is built and open-sourced.
	% The corresponding publicly available XJTU-SPS dataset includes four sub-datasets, involves 4 types of sub-tasks, 17 types of faults, over 700,000 timestamps.
	% The experimental results demonstrate that SpaceHMchat achieves excellent performance across 23 metrics, such as 100\% conclusion accuracy in logical reasoning of work condition recognition, over 99\% success rate in anomaly detection tool invocation, over 90\% precision in fault localization, and knowledge base search time under 3 minutes in maintenance decision-making.
\end{abstract}

\end{frontmatter}

% \linenumbers

\section{Introduction}
\label{Introduction}

With the rapid development of space technology in recent years, the number of spacecraft has increased explosively. 
Over thirty SMC are planned for deployment globally. 
Notable examples include Starlink project of SpaceX, which plans to deploy approximately 42,000 satellites; 
GW constellation project of China, which plans to deploy about 13,000 satellites; 
and the OneWeb constellation of the United Kingdom, which aims to deploy 6,372 satellites.
The future will undoubtedly be dominated by SMC. 
While the deployment of vast numbers of spacecraft significantly enhances global capabilities in communications, navigation, and remote sensing \cite{peters2018global, min2024lost}, it also introduces severe challenges, particularly regarding operational safety and maintenance.
Statistics indicate that the on-orbit anomaly rate for satellites exceeds 85\%, and over 45\% of satellites fail to reach their designed lifespan. 
Among all subsystems, the SPS is one of the most frequent sources of anomalies, accounting for up to 44\% of cases \cite{TAFAZOLI2009195, 9362601, wang2022case}. 
In particular, 45\% of SPS failures lead to mission failures, and approximately 80\% of failures have a significant impact on spacecraft missions \cite{TAFAZOLI2009195}.
In fact, 91\% of anomalies can be detected and resolved by ground health management teams, with only 9\% resulting in catastrophic failures \cite{8880888}.
Therefore, aerospace telemetry and command centers are equipped with teams of health management to monitor telemetry signals, execute the AIL HM process, and send various commands to spacecraft, such as mission scheduling, repair sequences, redundancy activation, parameter reconstruction, isolation restarts, etc.

The conventional AIL HM process includes the following steps \cite{juviken2025optimizing, o2011fault, 9172768} (as shown in Figure \ref{Fig_SpaceHMchat: a Human-AI collaboration framework for AIL HM, guided by the AUC principle}):
\textbf{(a) Work Condition Recognition:} 
To identify the system's operational status, the process begins with work condition recognition, establishing a health baseline and serving mission scheduling, attitude and orbit control, and anomaly detection.
Based on well-defined logic rules, decision trees, etc., space mission operators monitor telemetry and calculate relevant indicators to determine the system's work condition.
\textbf{(b) Anomaly Detection:} 
Based on advanced anomaly detection algorithms (statistical, machine learning, deep learning, etc.), telemetry data analysts debug and deploy anomaly detection algorithms, set thresholds, and monitor system data.
\textbf{(c) Fault Localization:} 
If the anomaly detection indicates an abnormal state, the process enters the fault localization phase, which supports subsequent root cause analysis and maintenance decision-making.
Based on historical fault cases and monitoring signals, senior subsystem experts continuously research and summarize to accumulate experience, analyze data to identify fault types.
\textbf{(d) Maintenance Decision-making:} 
Finally, as the last step in AIL HM, root cause analysis, risk assessment, and maintenance strategy are established.
On-orbit response teams consult technical documents, maintenance manuals, historical analysis reports, design documentation, and other relevant materials to formulate maintenance plans.

However, the advent of the SMC era poses significant challenges to the conventional pattern of AIL HM \cite{aerospace7090133, BENLARBI20213598}:
\textbf{(1) Manpower Cost Explosion:} 
In the SMC era, the number of spacecraft increases dramatically, leading to a surge in health management tasks.
However, maintaining hundreds of 7$\times$24-hour mission response teams staffed by senior experts is unattainable in the SMC era. 
This necessitates an optimized allocation of personnel, allowing experts to focus on more complex and high-value tasks, while routine tasks are handled by assistants.
\textbf{(2) Textual Information Deluge:} 
Current data analysis algorithms primarily focus on time-series telemetry sensor signals, neglecting another crucial information carrier.
The vast accumulation of design manuals, maintenance handbooks, maintenance logs, and technical documents in the SMC are critical sources of information.
However, these documents are voluminous and complex, relying on experts for reviewing and understanding, which is inefficient and error-prone.
\textbf{(3) Operational Complexity:} 
As the number of spacecraft types, models, and functionalities increases, the operational complexity also rises. 
For instance, the number of logical rules and indicators that need to be managed and memorized continues to grow, surpassing human capabilities.
Diagnosing certain intricate faults can take response teams days or even months, sometimes requiring tracing to original design and component selection documents.

Recent advancements in large language models (LLM) have demonstrated potential in handling the aforementioned challenges.
Alternative methods or auxiliary tools have been developed across various fields.
Notable advances are particularly rapid in the biomedical field:
For instance, using fine-tuning techniques, Lu et al. \cite{lu2024multimodal} proposed an interactive vision-language AI copilot that can flexibly handle both visual and natural language inputs for human pathology.
And based on retrieval-augmented generation (RAG) technology, Gaber et al. \cite{gaber2025evaluating} utilized LLMs to support clinical decision-making for triage, referral, and diagnosis.
Another widely adopted application of LLM is in the chemical and materials field:
Polak et al. \cite{polak2024extracting} used prompt engineering to develop an LLM-based approach, which extracts materials data from research papers.
Zhang et al. \cite{RN251} utilized LLMs to accelerate organic chemistry synthesis, exploring a suitable ligand and solvent within only 15 runs, achieving an isolated yield of 67\%.
Ruan et al. \cite{ruan2024automatic} developed specialized LLM agents capable of information extraction, substrate scope and condition screening, reaction kinetics study, reaction condition optimization, reaction scale-up, and product purification.
Meanwhile, there are also researches in the manufacturing and control fields:
Xu et al. \cite{XU2026111787} utilized an LLM-guided simulated annealing algorithm to optimize the selection and sequence of aero-engine rotor blades, generating interpretable assembly decision paths.
Aghaee et al. \cite{AGHAEE2026111259} leveraged the logical reasoning abilities of LLMs to serve the autonomous control systems of unmanned aerial vehicles, enhancing their decision-making capabilities in the presence of environmental disturbances and model uncertainties.
Zheng et al. \cite{ZHENG2025110548} proposed a LLM-based symbolic regression method to enhance the airfoil parameterization method.

\begin{figure}[htbp]
	\centering
	\includegraphics[page=1, scale = 1.0, trim=0 165 0 0, clip]{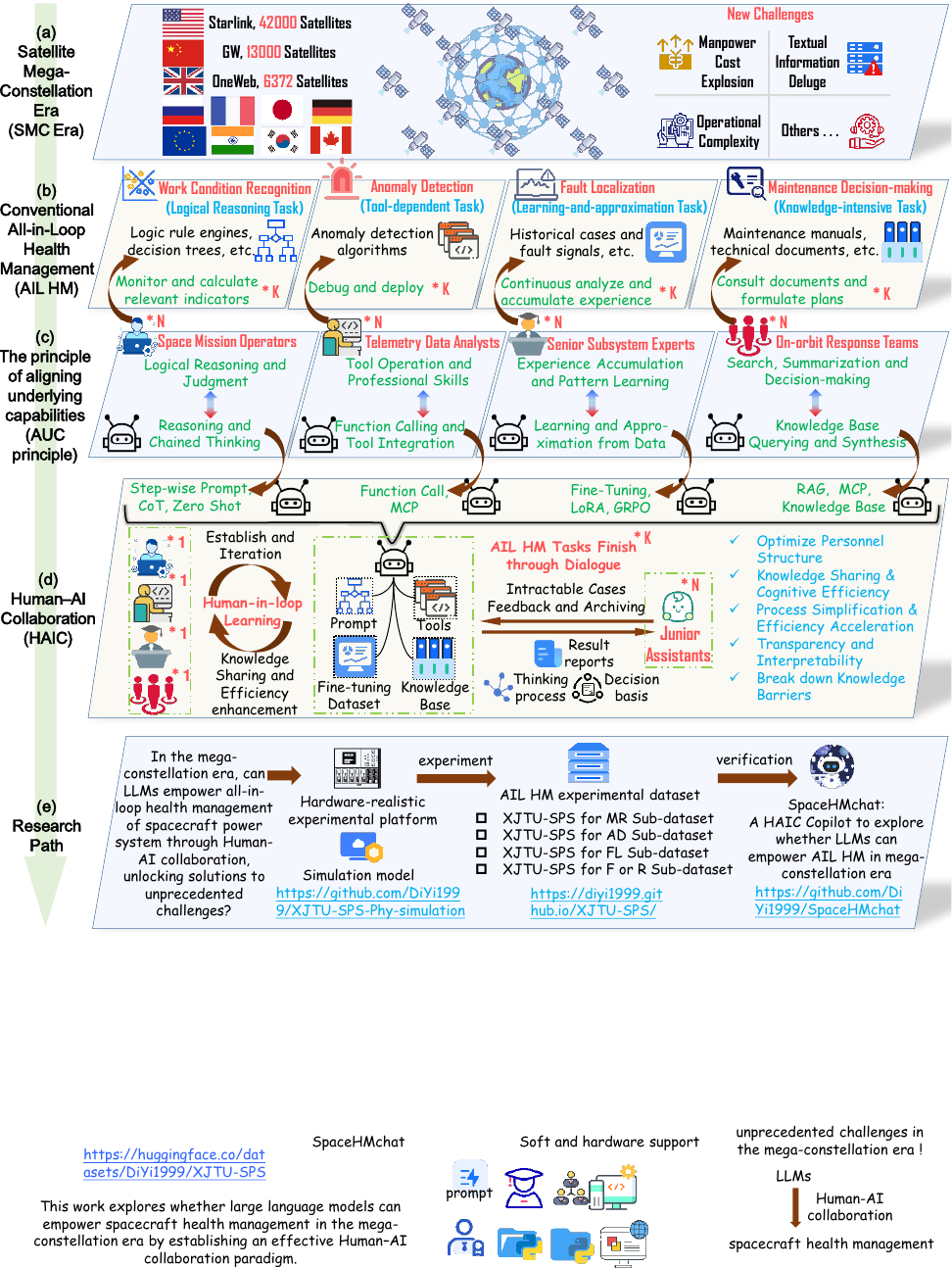}
	\caption{SpaceHMchat: a Human-AI collaboration framework for AIL HM, guided by the AUC principle. 
	(a) The future belongs to the SMC era, in which the explosive growth in spacecraft numbers introduces challenges, such as manpower cost explosion, textual information deluge, and operational complexity.
	(b) AIL HM encompasses work condition recognition, anomaly detection, fault localization, and maintenance decision-making, which are respectively performed by space mission operators, telemetry data analysts, senior subsystem experts, and on-orbit response teams.
	(c) This work proposes the AUC principle as the design principle of SpaceHMchat. 
	The core idea is to first identify the intrinsic nature of a given subtask, such as recognizing anomaly detection as a tool dependent task, then determine the underlying human capabilities required to accomplish it, such as tool operation and professional skills, and finally align these human capabilities with the corresponding capabilities of LLMs, such as function calling and tool integration, on the basis of which a HAIC framework is developed.
	(d) The developed HAIC framework, SpaceHMchat, enables human-in-loop learning and interaction with a low usage barrier, while delivering high quality and efficiency AIL HM, allowing senior experts to focus on more challenging tasks and the iterative improvement of SpaceHMchat, while routine tasks are rapidly and automatically executed by the copilot under assistant commands.
	(e) To validate the effectiveness and necessity of SpaceHMchat, both simulation models and a hardware-realistic fault injection experimental platform are established, resulting in the AIL HM dataset covering four subtasks. And all resources are open-sourced.
	}
	\label{Fig_SpaceHMchat: a Human-AI collaboration framework for AIL HM, guided by the AUC principle}
\end{figure}

Different from other fields, given the importance of space missions and the huge economic risks, sometimes also involving factors such as national defense and cybersecurity, the health management of SPS has attributes such as high reliability requirements, non-transferable responsibility, and absolute human control authority.
Thus, it excludes design of fully autonomous agent and instead emphasizes the HAIC framework as the primary design principle \cite{THANGAVEL2024100960, Bensch2025, aiaa2025Kerianne}.
Based on LLM technology, this work deploys a HAIC framework called SpaceHMchat to address the challenges faced by conventional pattern of SPS health management in the SMC era.
As shown in Figure \ref{Fig_SpaceHMchat: a Human-AI collaboration framework for AIL HM, guided by the AUC principle}, we propose the AUC principle to guide the construction of SpaceHMchat, which invokes the following viewpoints:
\textbf{(a) Work Condition Recognition:}
This stage is essentially a logical reasoning task,
the reasoning and chained thinking capabilities of LLMs can align with the logical reasoning and judgment abilities of humans.
Through techniques such as prompt engineering, chain-of-thought (CoT), and step-wise prompting, SpaceHMchat can apply rigorous logic rules and decision trees for work condition recognition.
\textbf{(b) Anomaly Detection:}
This stage is essentially a tool-dependent task,
the function calling and tool integration capabilities of LLMs can align with the tool operation and professional skills of humans.
Through techniques such as function calling and model context protocol (MCP), SpaceHMchat can invoke advanced anomaly detection algorithms.
\textbf{(c) Fault Localization:}
This stage is essentially a learning-and-approximation task,
the learning and approximation from data capabilities of LLMs can align with the experience accumulation and pattern learning abilities of humans.
Through techniques such as fine-tuning, LoRA (Low-Rank Adaptation), SFT (Supervised Fine-Tuning), and GRPO (Group Relative Policy Optimization), SpaceHMchat can learn to identify fault types from historical monitoring data, becoming an expert LLM with fault localization capabilities.
\textbf{(d) Maintenance Decision-making:}
This stage is essentially a knowledge-intensive task,
the knowledge base querying and synthesis capabilities of LLMs can align with the search, summarization, and decision-making abilities of humans.
Through techniques such as RAG, MCP, and knowledge bases, SpaceHMchat can efficiently review and understand vast amounts of textual information, assisting in root cause analysis, risk assessment, and maintenance decision-making.

% This work attempts to explore:
% \textit{Can we develop a HAIC framework, empowered by LLMs, to adapt the AIL HM of SPS in the SMC era, unlocking solutions to unprecedented challenges?}
As shown in Figure \ref{Fig_SpaceHMchat: a Human-AI collaboration framework for AIL HM, guided by the AUC principle}, SpaceHMchat, as a HAIC framework, achieves process simplification and efficiency acceleration while breaking down knowledge barriers, enabling junior assistants to automatically and rapidly complete most routine tasks through a conversational interface.
Furthermore, SpaceHMchat will optimize the personnel structure, allowing senior experts to intervene only in challenging tasks while focusing their efforts on iterating the tools and knowledge base of SpaceHMchat, thereby eliminating the need to allocate hundreds of expert teams for SMC.
And this HAIC framework ensures knowledge sharing and cognitive efficiency among experts, AI, and assistants, enabling efficient information retrieval and  achieving human-in-the-loop learning, accelerating the learning curve of assistants and experts.
Meanwhile, the results of SpaceHMchat include result reports, thinking processes, and decision basis, ensuring transparency and interpretability.
The main contributions of this work can be summarized as:
\begin{itemize}
	\item 
	To address the challenges encountered by conventional pattern of SPS health management in the SMC era, we propose the AUC principle and develop SpaceHMchat, an HAIC framework.
	% Based on techniques such as function calling, RAG, MCP, prompt engineering, and fine-tuning, SpaceHMchat can achieve interactive, low-barrier, and efficient health management, including work condition recognition, anomaly detection, fault localization, and maintenance decision-making for SPS.
	SpaceHMchat enables AIL HM, including work condition recognition, anomaly detection, fault localization, and maintenance decision-making.
	% Its advantages include personnel structure optimization, knowledge sharing and cognitive efficiency, process simplification and efficiency acceleration, transparency and interpretability, and the breakdown of knowledge barriers.
	The experimental results demonstrate that SpaceHMchat achieves excellent performance across 23 quantitative metrics, such as 100\% conclusion accuracy in logical reasoning of work condition recognition, over 99\% success rate in anomaly detection tool invocation, over 90\% precision in fault localization, and knowledge base search time under 3 minutes in maintenance decision-making.
	Its code is available at \url{https://github.com/DiYi1999/SpaceHMchat}.
	\item 
	To validate the potential of HAIC in the SMC era and verify the performance of SpaceHMchat, we establish a validation system encompassing simulation models, a hardware-realistic experimental platform, and an open-source dataset.
	Our hardware-realistic experimental platform fully replicates the real SPS, yielding the XJTU-SPS dataset, which contains four sub-datasets, involving 4 types of AIL HM sub-tasks, 6 data collection periods, 6 work conditions, and 17 types of faults, with over 700,000 timestamps.
	% Due to the strict confidentiality requirements in the aerospace field, simulation and data resources are scarce. 
	To the best of our knowledge, the XJTU-SPS dataset is the first publicly available AIL HM dataset for SPS.
	% The experimental results based on this dataset demonstrate the effectiveness of SpaceHMchat across 23 metrics, proving its potential and necessity in the SMC era.
	The details and download link of the simulation model and XJTU-SPS dataset are available at \url{https://diyi1999.github.io/XJTU-SPS/}.
\end{itemize}

\section{Methods}
\label{Methods}

\begin{figure}[htbp]
	\centering
	\includegraphics[page=2, scale = 1.0, trim=0 300 0 0, clip]{Draw.pdf}
	\caption{Flow diagram of our method.
		\textbf{(a) Work Condition Recognition:} 
		\textit{Question:} 
		Inquiries about work conditions;
		\textit{Main technique:} 
		Prompt Engineering, CoT; 
		\textit{Core idea:} 
		LLM reasons logically in a chain of thought, calculates indicators, and makes decisions about work conditions;
		\textit{Answer:} 
		The reasoning chain and conclusions about work conditions.
		\textbf{(b) Anomaly Detection:} 
		\textit{Question:} 
		Any requests related to inquiries, settings, and invoking programs for anomaly detection;
		\textit{Main technique:} 
		Function Calling, MCP;
		\textit{Core idea:} 
		SpaceHMchat understands user intent and invokes appropriate tools to advance anomaly detection;
		\textit{Answer:} 
		The results of tool execution, such as detected anomaly segments and visualizations.
		\textbf{(c) Fault Localization:} 
		\textit{Question:} 
		Questions about fault types;
		\textit{Main technique:} 
		Fine-Tuning, LoRA; 
		\textit{Core idea:} 
		SpaceHMchat learns from historical data to become an expert LLM, capable of judging fault types like human experts;
		\textit{Answer:} 
		The conclusions about fault types and the reason for judgment.
		\textbf{(d) Maintenance Decision-making:} 
		\textit{Question:} 
		Questions about root cause, risk assessment, and maintenance decision;
		\textit{Main technique:} 
		RAG, MCP, Knowledge Base;
		\textit{Core idea:} 
		SpaceHMchat efficiently reviews and understands vast amounts of professional knowledge base, assisting in risk assessment and maintenance decision-making;
		\textit{Answer:} 
		The possible root causes, similar cases, risk levels, and maintenance recommendations.
	}
	\label{Fig_Flow diagram of our method}
\end{figure}

\subsection{Work Condition Recognition}
\label{Work Condition Recognition}

As shown in Figure \ref{Fig_Flow diagram of our method}(a), this stage involves logical reasoning tasks, based on prompt engineering, CoT step-wise prompting, zero/few shot techniques, the development process of the work condition recognition part of SpaceHMchat is as follows:
\begin{enumerate}[1)]
	\item 
	\textbf{Build decision trees:}
	Current LLM technology excels at logical reasoning and metric calculation, but it has not yet achieved consistently reliable extraction of work condition decision rules from design documents.
	While this may become feasible in the future, the present work recommends that human designers pre-define the decision tree, which only needs to be performed once during the development phase.
	\item 
	\textbf{Design Step-wise Prompt:}
	Through prompt engineering techniques, the decision tree is perfectly transformed into LLM prompts.
	As shown in Figure \ref{Fig_Flow diagram of our method}(a), the core of this transformation is to convert the branches of the decision tree into step-wise if-else statements, and guide the LLM to reason in a CoT manner using ``Let's think step by step" or by providing a clear answer format.
	A referential template is as follows:
	\begin{tcolorbox}[colback=gray!20, colframe=gray!20, boxrule=0pt, width=0.98\textwidth, fontupper=\scriptsize\sffamily]
		``````

		\#\# \textcolor{red}{Instruction}: You are an expert of \textcolor{red}{[...]}, the following content is the analysis steps for executing work condition recognition tasks. If \textcolor{red}{[...]} Please \textcolor{red}{[...]}

		\#\# \textcolor{red}{Background}: The data is collected from \textcolor{red}{[...]}. The system is composed of \textcolor{red}{[...]}

		\#\# \textcolor{red}{Analysis Method}: Please follow the steps below to analyze:

		*** Step 1: Confirm whether the solar panel voltage is greater than 1V? If yes, record as Sunlit Area, move to Step 2; if no, record as Shadow Area, move to Step 6;
		
		*** Step 2: ...;

		...

		*** Step 6: Confirm whether the load current of each path is greater than 0.5A? If yes, record as Task \textless N \textgreater is being executed, output the work condition as [Discharge, Shadow Area, Task \textless N \textgreater]; if no, record as no task, output the work condition as [Idle, Shadow Area, No Task], move to Step 7;

		*** Step 7: ...;

		\#\# \textcolor{red}{Answer Format}: Please answer in the following format, replace the content in [] and \textless \textgreater
		:
		``
		Step 1: Confirm [\textit{matter to be confirmed}], [\textit{reasoning process of confirmation}], the answer is [\textit{yes/no}], record as [\textit{Sunlit Area/Shadow Area/Move to other steps}]; 
		Step 2: ...; ...
		Step 6: Confirm [\textit{matter to be confirmed}], [\textit{reasoning process of confirmation}], the answer is [\textit{yes/no}], record as [\textit{Task \textless N \textgreater/No Task/Move to other steps}]; 
		Step 7: ...; 
		In summary, the current work condition of the spacecraft is: [\textit{output work condition}].
		"

		\#\# \textcolor{red}{Note}: 1. If ...; 2. If ...

		\#\# \textcolor{red}{Answer Example}: Meanwhile, here are some examples of correct answers for your reference: "\textless classic sample example 1 \textgreater", "\textless classic sample example 2 \textgreater", "\textless classic sample example 3 \textgreater"...

		"""
	\end{tcolorbox}
	In most cases, zero-shot is sufficient to meet the requirements. 
	If some samples are available, they can be added to the prompt in the ``Answer Example''. 
	This few-shot approach can be beneficial.
	A complete example of the work condition recognition prompt and its corresponding decision tree is provided in \ref{Appendix: Prompt Example of Work Condition Recognition}.
	
	\item 
	\textbf{Get data and pack prompt:}
	The designed step-wise prompt is used as the system prompt, along with user question, history contexts, and data segment information, to be packed and input to LLM.
	
	\item
	\textbf{CoT Return:}
	LLM reasons logically in a CoT manner, calculates indicators, and makes decisions about work conditions.

\end{enumerate}
In this task, the space mission operator only needs to perform the decision tree formulation and prompt design once.
The prompt will then be saved in the backend of the HAIC framework and reused, allowing junior assistants to obtain results directly through the dialogue without understanding the specific principles and methods of work condition recognition, serving subsequent tasks.
And during this process, LLM ensures transparency and interpretability through its unique CoT feedback.
In fact, this can be seen as permanently teaching the expertise of space mission operators to LLMs, continuously optimizing and improving it during the version iterations of SpaceHMchat.

\subsection{Anomaly Detection}
\label{Anomaly Detection}

As shown in Figure \ref{Fig_Flow diagram of our method}(b), this stage involves tool-dependent tasks, based on function calling and MCP techniques, the development process of the anomaly detection part of SpaceHMchat is as follows:
\begin{enumerate}[1)]
	\item 
	\textbf{Construct Database, Model-set, Data-stream, Device-management, etc.:}
	The foundation includes: database such as historical monitoring data; model-set composed of algorithms such as statistics, machine learning, deep learning like MTGNN (Multivariate Time Graph Neural Network), GRU (Gated Recurrent Unit), TCN (Temporal Convolutional Network), Transformer; data-stream such as streaming reading interfaces for real-time telemetry data; device-management such as management of algorithm running environment, CPU, GPU and other resources.
	These foundations always have many mature open-source projects to refer to, such as CUDA \cite{cuda}, BasicTS \cite{shao2024exploring}, etc.
	\item 
	\textbf{Build Toolkit:}
	Although there are many mature MCPs, some important and frequently called tools are still recommended to be implemented through function calling to ensure calling stability and authorization control, including data load tool, model selection and parameter setting tool, model training tool, model inference tool, visualization tool, etc.
	The MCP is deployed for some supplementary and temporary functions, such as web queries.
	\item 
	\textbf{Message Pack:}
	Pack system prompt, user question, and tool description into a message pack, input to the LLM.
	\item
	\textbf{Function Call / MCP:}
	SpaceHMchat understands user intent and returns tool call information, such as calling anomaly detection method A, importing training weights B, performing anomaly detection on data segments C.
	\item
	\textbf{Tool Execution and Results Return:}
	Based on database and model-set, etc., the tool executes and returns the results.
\end{enumerate}
In this task, the telemetry data analyst only needs to be responsible for debugging, deploying, maintaining, and iterating anomaly detection algorithms.
They can focus on developing more advanced algorithms, which will improve the effectiveness of anomaly detection.
Junior assistants can directly command the corresponding tools to complete anomaly detection tasks through the dialogue interface, this process simplification accelerates detection efficiency.
Any questions during this process can be directly asked to SpaceHMchat.

\subsection{Fault Localization}
\label{Fault Localization}

As shown in Figure \ref{Fig_Flow diagram of our method}(c), this stage involves learning-and-approximation tasks, based on fine-tuning, LoRA, and other techniques, the development process of the fault localization part of SpaceHMchat is as follows:
\begin{enumerate}[1)]
	\item 
	\textbf{Build Q\&A Dataset:}
	The prerequisite for enabling LLM to learn from historical data and become an expert LLM is to convert time series monitoring data into a question and answer dataset.
	A referential template is as follows:
	\begin{tcolorbox}[colback=gray!20, colframe=gray!20, boxrule=0pt, width=0.98\textwidth, fontupper=\scriptsize\sffamily]
		``````

		\#\#\# \textcolor{red}{Instruction}: Based on the provided data, please identify which type of fault occurred. Possible fault types include: [BCR short circuit \textcolor{red}{...}].
    
		\#\#\# \textcolor{red}{Description}: This data was collected from \textcolor{red}{...}
		\hspace{1cm}  % 加空格
		\#\#\# \textcolor{red}{Data: ...}
		\hspace{1cm}  % 加空格
		\#\#\# \textcolor{red}{Answer format}: \textcolor{red}{...}
		\hspace{1cm}  % 加空格
		\#\#\# \textcolor{red}{Answer}: \textcolor{red}{...}

		"""
	\end{tcolorbox}
	The time series data is filled in the "Monitoring data" section, which can be directly filled in, or scaled, binned, or with special characters added \cite{wang2025chattime}, or features can be extracted and inserted.
	Specific filling methods can refer to those used by human senior experts, and practice demonstrates that excessive preprocessing such as scaling is not effective.
	A complete example of the Q\&A template and some referential feature candidates are provided in \ref{Appendix: Q&A Template for Fault Localization}.

	\item 
	\textbf{Fine-Tuning:}
	Through fine-tuning and LoRA techniques, the LLM is fine-tuned into an expert LLM, enabling it to learn from historical monitoring data and judge fault types.
	Due to the demands on LLM's understanding and reasoning capabilities to align with human senior experts, it is recommended to choose a larger base model for fine-tuning (greater than or equal to 14B), such as Qwen3-14B.
	
	\item 
	\textbf{Get data and pack message:}
	According to the user's inquiry, after importing the queried data segment, pack the user question, data segment, system prompt, and history contexts into a message pack, and input it to the expert LLM.
	
	\item
	\textbf{Return Result and Reason:}
	SpaceHMchat returns the conclusion about the fault type, the reason for judgment, and the thought process,
	such as \textit{"$<$ think $>$ … $<$ /think $>$   $<$ answer $>$ The type of fault in this data segment is …   The analysis is based on the fact that the voltage sensor monitoring data … $<$ /answer $>$"}

\end{enumerate}
In this task, senior subsystem experts focus on summarizing their expertise into the Q\&A dataset.
Through fine-tuning techniques, the LLM is transformed into an expert LLM, some already occurred or common faults do not require consulting senior subsystem experts, can be directly localized through SpaceHMchat.
Even when encountering some intractable faults, junior assistants can request expert assistance once and archive new knowledge, through continuous learning, SpaceHMchat is expected to surpass human experts in comprehensive capabilities.

\subsection{Maintenance Decision-making}
\label{Maintenance Decision-making}

As shown in Figure \ref{Fig_Flow diagram of our method}(d), this stage involves knowledge-intensive tasks, based on RAG, MCP, knowledge base, and other techniques, the development process of the maintenance decision-making part of SpaceHMchat is as follows:
\begin{enumerate}[1)]
	\item 
	\textbf{Build Knowledge Base:}
	The effectiveness of this task largely depends on the richness and expertise of the knowledge base.
	The knowledge base includes historical root cause analysis reports, design documentation, global spacecraft accident case studies, maintenance technical documentation, relevant papers and technical standards, archived expert consultations, etc.
	Given that LLMs can process and comprehend vast textual information at unprecedented speeds, the knowledge base scale can exceed human experts' capacity. 
	Any domain-specific documentation can be integrated, though the knowledge base require consistent updating.

	\item 
	\textbf{Query Vector Database:}
	RAG is recommended to ensure stable query quality and efficiency.
	It can process the textual information in the knowledge base through tokenization, embedding, vectorization, etc, and then store in a vector database.
	
	\item 	
	\textbf{Network Info Collection:}
	In addition to the knowledge base, to ensure the acquisition of real-time information and further expand information sources, it is recommended to integrate web query functionality through MCP. 
	This can save human experts a significant amount of time spent on manual queries.
	
	\item
	\textbf{Retrieve Top-k Passage:}
	Based on the user's inquiry, retrieve the top-k most relevant textual information from the vector database through vectorized matching.
	
	\item
	\textbf{Message Pack:}
	Pack system prompt, user question, retrieved top-k passage, and history contexts into a message pack, and input it to SpaceHMchat.
	A referential system prompt template is provided in \ref{Appendix: Prompt Example for Maintenance Decision-making}.

	\item
	\textbf{Results Return:}
	SpaceHMchat returns the root cause of the fault, similar cases, risk levels, maintenance suggestions, etc.

\end{enumerate}
In this task, on-orbit maintenance teams focus on the construction and maintenance of the knowledge base, ensuring absolute knowledge sharing and cognitive efficiency.
Not only can junior assistants complete routine maintenance decision-making tasks through querying the knowledge base, but even experts can also perform knowledge base queries and web queries through SpaceHMchat when encountering complex problems, which is more efficient, as LLMs can process massive volumes of textual information with high precision and speed, reducing repetitive work.
In fact, the ongoing process of document updates and knowledge base expansion also serves as a process of knowledge purification, filtration, and the establishment of unified cognition.

\section{Experiments}
\label{Experiments}

\subsection{XJTU-SPS Dataset}
\label{XJTU-SPS Dataset}

To evaluate the performance of SpaceHMchat, a hardware-realistic experimental platform that fully replicates the real SPS has been built.
This platform is able to perform fault injection, and the obtained XJTU-SPS dataset contains four sub-datasets for AIL HM, simulating telemetry data from a low Earth orbit SPS and monitoring 33 sensors under various working conditions and fault scenarios.
(1) XJTU-SPS for MR Sub-dataset (for Work Mode Recognition Task): Simulate working conditions such as Fast Charge, Shunt, Trickle Charge, Joint, Idle, and Discharge, etc.
(2) XJTU-SPS for AD Sub-dataset (for Anomaly Detection Task): Simulate the situations when various anomalies occur during the operation, exceeding 700,000 timestamps.
(3) XJTU-SPS for FL or FD Sub-dataset (for Fault Localization / Fault Diagnosis Task): Simulate 17 types of fault scenarios, including partial component or branch open circuit of SA, BCR short circuit, and Bus insulation breakdown, etc.
(4) XJTU-SPS for F or R Sub-dataset (for Forecasting or Reconstruction Task): Six sub-files simulate the data of 4, 18, 24, 34, 90, and 94 orbits around the Earth, respectively, with a sampling frequency of 1Hz.
As far as we know, it is the first publicly available AIL HM dataset in the field, dedicated to facilitate research in the energy sector and in the aerospace sector, where data resources are scarce due to stringent confidentiality constraints.
Meanwhile, a simulation model corresponding to this dataset has also been established, which is developed according to the design principles and working mechanisms of real SPS, capable of highly restoring the operating status and dynamic characteristics of real SPS, further supporting researchers in fields such as digital twins and physics-informed neural networks.
The detailed information of the XJTU-SPS dataset and simulation model of SPS can be found in \ref{Appendix: XJTU-SPS Dataset and Simulation Model}. Their open-source link is located at \url{https://diyi1999.github.io/XJTU-SPS/}.

\subsection{Results of Work Condition Recognition}
\label{Results of Work Condition Recognition}

\begin{figure}[htbp]
	\centering
	% \subfigure[The interaction results of work condition recognition task.]{
	% 	\label{Fig_The interaction results of work condition recognition task}
		
		\begin{minipage}[c]{0.69\textwidth}
			\begin{spacing}{1.2} % 行距倍率
				\begin{chat}

					\chatdate{SpaceHMchat}{}

					\chatpersonOne{
						During the time period $['2024\text{-}10\text{-}18$ $18\text{:}43\text{:}23'$, $'2024\text{-}10\text{-}18$ $18\text{:}44\text{:}15']$, what was the spacecraft's work condition? 
						Please execute the work condition recognition task.
					}

					\chatpersonTwo{
						\textcolor{blue}{\textbf{$<$ think $>$}}
						\\
						Okay, I now need to handle the user's work condition recognition task. First, the user has provided detailed sensor data, and \textcolor{red}{\textbf{I need to analyze it step by step according to the previously given steps.}}
						Step 1 is to confirm whether the solar panel voltage is greater than 1V. The user's data is 28.746V, which is obviously greater than 1V, so I record it as sunlit area and enter Step 2.
						Step 2 checks ...
						\\
						...
						Overall, the work condition should be CV charging, sunlit area, no task. Need to check for any anomalies, but based on the data, it seems normal.
						\\
						$<$ \textbackslash \textcolor{blue}{\textbf{think}} $>$
						\\
						\textcolor{blue}{\textbf{$<$ answer $>$}}
						\\
						\textcolor{red}{\textbf{Step 1}}: Confirm whether the solar panel voltage is greater than 1V; the solar panel voltage is 28.746V, the answer is yes, record as sunlit area, proceed to Step 2; 
						\textcolor{red}{\textbf{Step 2}}: Confirm whether the currents of each load are greater than 0.5A; the currents of loads 1/2/3 are all 0.0A, the answer is no, record as no task, proceed to Step 4; 
						\textcolor{red}{\textbf{Step 3}}: Not triggered, proceed to Step 4; 
						\textcolor{red}{\textbf{Step 4}}: Confirm whether the current values of each battery pack are less than -0.1A; the currents of battery packs 2/3/4 are -0.441A/-0.449A/-0.423A respectively, the answer is yes, proceed to Step 5; 
						\textcolor{red}{\textbf{Step 5}}: Confirm whether the battery pack voltage increase is greater than 0.05V; the battery pack voltages are all in the 16.765-16.769V range, the answer is no, record as CV charging, output operating state as $\llbracket$ CV charging, sunlit area, no task $\rrbracket$, proceed to Step 7; 
						\textcolor{red}{\textbf{Step 6}}: Not triggered; 
						\textcolor{red}{\textbf{Step 7}}: Check the observation values of each sensor: the solar panel voltage 28.746V is within the normal range of 0-32V, the BCR current 2.226A is within the allowable charging range of 0-3.5A, the battery pack current ..., the inspection process reasoning conforms to the characteristics of CV charging, recorded as inspection normal; 
						\textcolor{red}{\textbf{In summary, the spacecraft's work condition is: $\llbracket$ \textcolor{red}{\textbf{CV charging, sunlit area, no task}} $\rrbracket$.}}
						\\
						$<$ \textbackslash \textcolor{blue}{\textbf{answer}} $>$
					}

				\end{chat}
			\end{spacing}
		\end{minipage}
		\hfill
		\begin{minipage}[c]{0.30\textwidth}
			\begin{spacing}{1.2}
				\begin{chat}

					\chatdate{Base Model}{}

					\chatpersonOne{
						During the time period $['2024\text{-}10\text{-}18$ $18\text{:}43\text{:}23'$, $'2024\text{-}10\text{-}18$ $18\text{:}44\text{:}15']$, what was the spacecraft's work condition? 
						Please execute the work condition recognition task.
					}

					\chatpersonThree{
						To assess the spacecraft's work condition during the time period $['2024\text{-}10\text{-}18$ $18\text{:}43\text{:}23'$, $'2024\text{-}10\text{-}18$ $18\text{:}44\text{:}15']$, we perform a work condition recognition task based on telemetry data
						\\
						\textcolor{blue}{\textbf{Conclusion}}:
						At $'2024\text{-}10\text{-}18$ $18\text{:}43\text{:}23'$, the spacecraft was operating under \textcolor{red}{\textbf{nominal work conditions across all major subsystems}}. No precursors or degradations were observed in power, thermal, attitude control, communications, or computing systems.
						\\
						\textcolor{blue}{\textbf{Detailed Analysis}}:
						\\
						1. Solar Array Performance:
						- Voltage: 29.42 V (stable)
						- Current: 1.01 A (slightly decreasing from 1.0114 to 1.009 A)
						- Power: 29.7-29.8 W (consistent output).
						2. Thermal: ...
						\\
						...
					}

				\end{chat}
			\end{spacing}
		\end{minipage}
	% }

	% \vspace{-0.7cm} % 调整图片与图标题之间太宽的间距空白

	% \subfigure[The evaluation results of SpaceHMchat on work condition recognition task.]{
	% \label{Fig_The evaluation results of SpaceHMchat on work condition recognition task}
	% \begin{minipage}{0.98\textwidth}
	% 	% \begin{spacing}{1.2} % minipage行距倍率
	% 	% \centering

	% \end{minipage}	
	% }

	\caption{The interaction results of work condition recognition task.
			Our proposed SpaceHMchat's dialogue results are shown on the left, while the right side shows the response of a base model without any techniques mentioned in Figure \ref{Fig_Flow diagram of our method}.
			An example of AIL HM dialogue with complete information is shown in \ref{Appendix: Complete Chat Examples}.
	}
	\label{Fig_Results of SpaceHMchat on Work Condition Recognition Task}
	\end{figure}

The results of the work condition recognition task are shown in Figure \ref{Fig_Results of SpaceHMchat on Work Condition Recognition Task}.
The success of this practice demonstrates that LLMs possess sufficient reasoning and chained thinking capabilities to accurately execute binary tree-based work condition identification tasks with fully specified procedural steps.
Its advantages mainly lie in:
\textbf{(1).} 
The process is simplified, space mission operators no longer need to perform logical deductions or invoke any program, instead relying solely on asking questions.
\textbf{(2).} 
The results exhibit enhanced precision and interpretability, with the strictly structured reasoning chain ensuring stable and error-free outcomes while recording and displaying all steps for improved transparency and usability.
Regarding specific implementation techniques, the effectiveness of prompt engineering and CoT is satisfactory.
The base model (Qwen3-235B) provides conclusions directly, essentially is a form of generation without prior, unable to yield correct results.

\subsection{Results of Anomaly Detection}
\label{Results of Anomaly Detection}

\begin{figure}[htbp]
	\centering
	\begin{minipage}[c]{0.61\textwidth}
		\begin{spacing}{1.2} % 行距倍率
			\begin{chat}

				\chatdate{SpaceHMchat}{}
				% \chatdate{SpaceHMchat}{ (anomaly detection)}

				\chatpersonOne{
					Please execute the anomaly detection task, which algorithms are available?
				}

				\chatpersonTwo{
					\textcolor{red}{\textbf{Based on the outcomes of the tool executions, the available anomaly detection algorithms include:}} 
					$[$`MTGNN', `FourierGNN', `StemGNN', `GraphWaveNet', `GAT', `GCN', `GIN', `SGC', `Transformer', `Informer', `Autoformer', `GRU', `LSTM', `RNN', `TCN', `MLP', `DLinear', `PatchTST', ...$]$.
					\\
					\textcolor{red}{\textbf{More Information:}}
					1. Time Series Models: Transformer, ...;
					2. GNN based Models: MTGNN, ...;
					3. ...
					\\
					\textcolor{red}{\textbf{Suggestions for Selection:}}
					1. For multi-variate time series data, consider MTGNN, ...;
					2. For long-horizon forecasting tasks, consider Transformer, ...;
					3. ...
					\\
					\textcolor{red}{\textbf{Parameter Setting Tips:}}
					Most algorithms have default parameters set, users only need to ...
					You can specify a particular algorithm to check its default parameters and configuration requirements.
				}

				\chatpersonOne{
					I want to use MTGNN, are there any parameters I need to set?
				}

				\chatpersonTwo{
					\textcolor{red}{\textbf{Based on the outcomes of the tool executions, the parameters that need to be set for MTGNN are:}}
					1. MTGNN\_layer\_norm\_affline (default: True): Whether to perform element-wise affine operation in layer normalization;
					2. MTGNN\_layers (default: 3): Number of GC + TC layers;
					3. MTGNN\_node\_embedding (default: 192): Dimension of node embedding;
					4. ...
					\\
					\textcolor{red}{\textbf{More Info:}}
					\\
					\textcolor{red}{\textbf{Suggestions for Core Parameters:}}
					1. ...; 2. ...
					\\
					\textcolor{red}{\textbf{Graph Structure Related Parameters:}}
					1. ...; 2. ...
					\\
					...
					\\
					\textcolor{red}{\textbf{Special Considerations:}}
					1. ...; 2. ...
				}

				\chatpersonOne{
					Change these parameters of MTGNN: set batch\_size to 64, set patience to 5, ...
				}

				\chatpersonTwo{
					\textcolor{red}{\textbf{The specified parameters of MTGNN have been successfully set/modified.}}
					\\
					\textcolor{red}{\textbf{Additional Info:}}
					1. A larger batch\_size can speed up training but requires more memory/VRAM, ...;
					2. Patience=5 means that if the validation loss does not improve for 5 consecutive epochs, ...;
					3. ...
				}

				\chatpersonOne{
					According to the set parameters, train the MTGNN model using the data from \textit{/data/.../XJTU-SPS Dataset/XJTU-SPS for AD/XJTU-SPS for AD\_Train.csv}.
				}

				\chatpersonTwo{
					\textcolor{red}{\textbf{The training is successfully complete and the weights saved to: \textit{/data/.../ckpt}.}}
				}

				\chatpersonOne{
					Using the previously trained anomaly detection algorithm MTGNN, perform anomaly detection on the data from \textit{/data/.../XJTU-SPS Dataset/XJTU-SPS for AD/XJTU-SPS for AD\_Test.csv}.
				}

				\chatpersonTwo{
					\textcolor{red}{\textbf{The anomaly detection is successfully complete, the detection results are as follows:}}
					\\
					\textcolor{red}{\textbf{1. Detected Anomaly Timestamp Intervals:}}
					\\
					$['2024\text{-}10\text{-}18 \quad  18\text{:}43\text{:}23', '2024\text{-}10\text{-}18 \quad  18\text{:}44\text{:}15']$,
					\\
					$['2024\text{-}10\text{-}18 \quad  18\text{:}44\text{:}17', '2024\text{-}10\text{-}18 \quad  18\text{:}45\text{:}18']$,
					\\
					$['2024\text{-}10\text{-}18 \quad  18\text{:}47\text{:}20']$,
					\\
					... ... (preview folded, download TXT file for complete results)
					\\
					\includegraphics[height=1em]{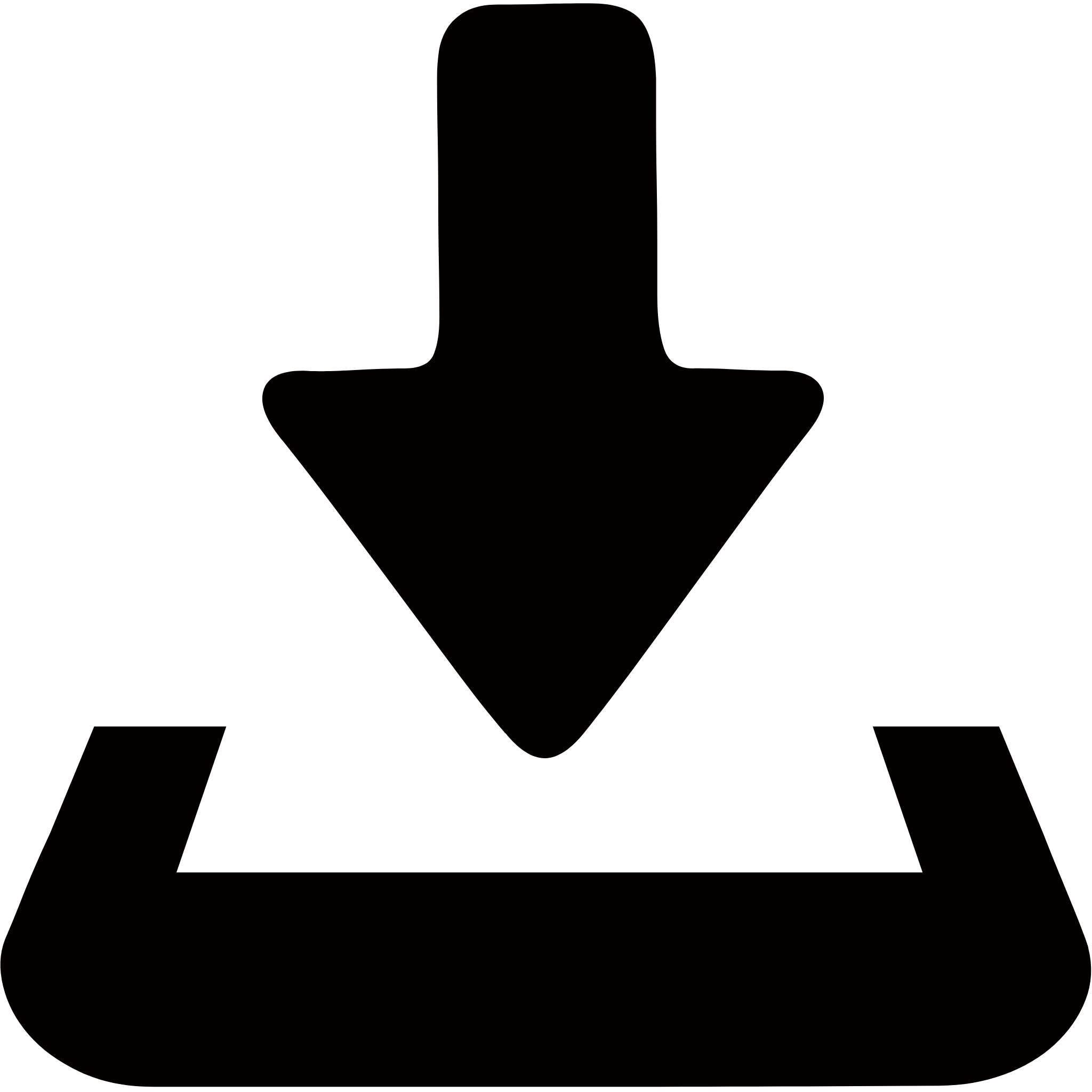}
					\textcolor{blue}{Complete Results Download (TXT)} (Single-element lists indicate point anomalies, double-element lists indicate continuous anomalies.)
					\\
					\textcolor{red}{\textbf{2. Visualization of Anomaly Detection Results:}}
					\\
					\hspace*{5mm}
					\includegraphics[width=0.65\textwidth, height=0.12\textwidth]{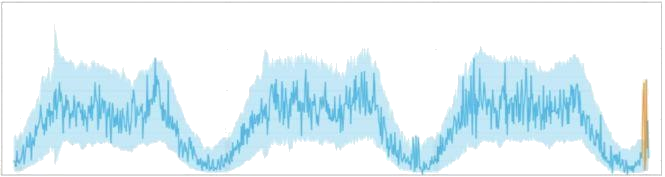}
					\\
					\hspace*{5mm}
					\includegraphics[height=1em]{download.png}
					\textcolor{blue}{Complete Visualization Download (PNG) ...}
					\\
					\textcolor{red}{\textbf{3. Overview of Anomaly Detection Result Information:}}
					\\
					- Anomaly ratio: 0.1488\%;
					\\
					- Detection threshold: 0.0156;
					\\
					...
					\\
					\textcolor{red}{\textbf{4. Additional Notes:}}
					\\
					- The complete anomaly detection report file is saved in \textit{/DiYi\_LLM\_Project/.../report}.
					\\
					- If you find the anomaly detection results unsatisfactory, here are some suggestions ...
				}
						
			\end{chat}
		\end{spacing}
	\end{minipage}
	\hfill
	\begin{minipage}[c]{0.38\textwidth}
		\begin{spacing}{1.2} % 行距倍率
			\begin{chat}

				\chatdate{Base Model}{}
				% \chatdate{Base Model}{ (anomaly detection)}

				\chatpersonOne{
					Please execute the anomaly detection task, which algorithms are available?
				}

				\chatpersonThree{
					\textcolor{red}{\textbf{I cannot directly access your local file system or the contents of your OneDrive directory, so I can't see the specific algorithms you have there.}}
					However, I can provide a general overview of common anomaly detection algorithms and frameworks that are widely used in the field.
					% For anomaly detection tasks, there are numerous algorithms available, and the choice depends on your data type (e.g., numerical, categorical, time-series), data size, whether anomalies are labeled, and the specific domain. Here's a structured overview of common algorithms and frameworks:
					% \\
					% \textbf{1. Unsupervised Anomaly Detection (No labeled anomalies required)}
					% \\
					% \textit{Classic Algorithms:  }
					% Isolation Forest: ...
					% One-Class SVM: ...
					% ...
					% \\
					% \textit{Ensemble/Advanced Methods:  }
					% ..., 
					% ...
					% \\
					% \textbf{2. Supervised Anomaly Detection (Requires labeled anomalies)}
					% \\
					% XGBoost / LightGBM: ...
					% ...
					% \\
					% \textbf{3. ...}
					% % \textcolor{red}{\textbf{3. Specialized Use Cases}}
					% % ...
					% % \textcolor{red}{\textbf{4. ...}}
					% \\
					% % For tabular data with no labels → Isolation Forest or PyOD.
					% % For time-series → TODS or LSTM Autoencoders.
					% % For fraud detection (labeled) → XGBoost with SMOTE.
					\\
					...
					\\
					Let me know your use case, and I'll suggest a tailored approach! 
				}

				\chatpersonOne{
					I want to use MTGNN, are there any parameters I need to set?
				}

				\chatpersonThree{
					MTGNN (Multi-Time Graph Neural Network) is a powerful model for spatio-temporal forecasting (e.g., traffic prediction, weather forecasting). To configure it effectively, you need to set both model architecture parameters and training hyperparameters. 
					Below is a structured breakdown of key parameters:
					...
				}

				\chatpersonOne{
					Change these parameters of MTGNN: set batch\_size to 64, set patience to 5, ...
				}

				\chatpersonThree{
					\textcolor{red}{\textbf{I can not directly modify code or configuration files on your system.}} 
					However, I can guide you on how to do it.
					Depending on your setup (e.g., training script, configuration file, or command-line arguments), here are a few common ways this might look:
					\\
					\textbf{Option 1: If using command-line arguments}
					% \\
					...
					\\
					\textbf{Option 2: If modifying a config dictionary in Python}
					% \\
					...
				}

				\chatpersonOne{
					According to the set parameters, train the MTGNN model using the data from \textit{/data/.../XJTU-SPS Dataset/XJTU-SPS for AD/XJTU-SPS for AD\_Train.csv}.
				}

				\chatpersonThree{
					The request involves accessing local files, executing algorithms on local hardware, and saving model weights locally. \textcolor{red}{\textbf{However, I do not have access to your local file system, hardware, or codebase, and none of the tools provided allow interaction with local environments.}}
					I can only perform internet searches to help guide you: ...
				}

				\chatpersonOne{
					Using the previously trained anomaly detection algorithm MTGNN, perform anomaly detection on the data from \textit{/data/.../XJTU-SPS Dataset/XJTU-SPS for AD/XJTU-SPS for AD\_Test.csv}.
				}

				\chatpersonThree{
					The request involves accessing local model weights, executing inference on test data, and saving results/visualizations to your local system. \textcolor{red}{\textbf{However, I cannot interact with your local files, hardware, or execute code directly. None of the provided tools enable access to your environment.}}
					Let me know the specific guidance you need, ...
				}
						
			\end{chat}
		\end{spacing}
	\end{minipage}

	% \vspace{-0.7cm} % 调整图片与图标题之间太宽的间距空白

	\caption{Results of SpaceHMchat on Anomaly Detection Task.
			Our proposed SpaceHMchat's dialogue results are shown on the left, while the right side shows the response of a base model without any techniques mentioned in Figure \ref{Fig_Flow diagram of our method}.
			An example of AIL HM dialogue with complete information is shown in \ref{Appendix: Complete Chat Examples}.
	}
	\label{Fig_Results of SpaceHMchat on Anomaly Detection Task}
	\end{figure}

The interaction results of the anomaly detection task are shown in Figure \ref{Fig_Results of SpaceHMchat on Anomaly Detection Task}.
The success of this practice demonstrates that the function calling and tool integration capabilities of LLMs are sufficient to transform conventional anomaly detection procedures into a conversational interaction paradigm. 
Its advantages mainly lie in:
\textbf{(1).} 
It eliminates programming-related technical steps, significantly lowering the skill barrier. 
Data analysts can focus on maintaining the algorithm library, while the specific detection workflow is executed by No-Code Data Operator and SpaceHMchat.
\textbf{(2).}
It provides constructive guidance, including recommendations for algorithm selection, parameter configuration, and result interpretation, etc. 
Regarding specific implementation techniques, the efficacy of function calling and MCP demonstrates satisfactory performance. 
The base model (Qwen3-235B) is constrained by its inability to access to local files and algorithm libraries, failing to execute user-specified operations.

\subsection{Results of Fault Localization}
\label{Results of Fault Localization}

\begin{figure}[htbp]
	\centering

	% \subfigure[The comparison of SpaceHMchat and its base model on fault localization task.]{
	% 	\label{Fig_Comparison of SpaceHMchat and Base Model on Fault Localization Task}

		\begin{minipage}[c]{0.63\textwidth}
			\begin{spacing}{1.2} % 行距倍率
				\begin{chat}

					\chatdate{SpaceHMchat}{}

					\chatpersonOne{
						I found that you detected a large area of anomaly near $'2024\text{-}12\text{-}05 \quad  22\text{:}12\text{:}50'$. Please execute the fault localization task, tell me what type of fault has occurred around $'2024\text{-}12\text{-}05 \quad  22\text{:}12\text{:}50'$?
					}

					\chatpersonTwo{
						\textcolor{red}{\textbf{The fault type for this data segment is an open circuit in a battery pack.}}
						\\
						\textcolor{red}{\textbf{The analysis is based on the fact that}} the solar panels are in eclipse (voltage 0.24 V, current 0 A), indicating reliance on the battery packs for power. At this time the total load current reaches 2.9 A, the bus voltage is 16.4 V and the power is stable (16.8 W), confirming that energy comes from the battery packs. Under normal conditions, the three battery packs should output in parallel, but the current sensor data for battery pack 2 is 0, while battery packs 3 and 4 have discharge currents of 0.34 A and 0.35 A respectively (within the -0.7 to +0.5 A range). According to design specifications, at the same voltage the battery packs should jointly share the load, but battery pack 2 has no current output and a stable temperature (26.6 °C), indicating an internal open-circuit fault that prevents it from supplying in parallel with the other packs, resulting in only battery packs 3 and 4 supplying the load, which is consistent with characteristics of a battery pack open-circuit fault.
					}
				\end{chat}
				
				% \vspace{-6pt} % 调整图片与图标题之间太宽的间距空白

			\end{spacing}
		\end{minipage}
		\hfill
		\begin{minipage}[c]{0.36\textwidth}
			\begin{spacing}{1.2} % 行距倍率
				\begin{chat}

					\chatdate{Base Model}{}

					\chatpersonOne{
						I found that you detected a large area of anomaly near $'2024\text{-}12\text{-}05 \quad  22\text{:}12\text{:}50'$. Please execute the fault localization task, tell me what type of fault has occurred around $'2024\text{-}12\text{-}05 \quad  22\text{:}12\text{:}50'$?
					}

					\chatpersonThree{
						Okay, let me try to figure out what's going on here. So, the user provided data from a spacecraft power system, and they think there's a fault. The possible issues are listed, like solar panel components open or shorted, battery charger controller (BCR) issues, etc.
						First, I need to look at ...
						\\
						So the faults detected are: solar panel part open/short, BCR open or short, battery aging, bus open or short.
					}
				\end{chat}
				
				% \vspace{-6pt} % 调整图片与图标题之间太宽的间距空白

			\end{spacing}

		\end{minipage}
	% }

	% \vspace{-0.7cm} % 调整图片与图标题之间太宽的间距空白

	\caption{Results of SpaceHMchat on Fault Localization Task.
			Our proposed SpaceHMchat's dialogue results are shown on the left, while the right side shows the response of a base model without any techniques mentioned in Figure \ref{Fig_Flow diagram of our method}.
			An example of AIL HM dialogue with complete information is shown in \ref{Appendix: Complete Chat Examples}.
	}
	\label{Fig_Results of SpaceHMchat on Fault Localization Task}
\end{figure}

The results of the fault localization task are shown in Figure \ref{Fig_Results of SpaceHMchat on Fault Localization Task}.
The success of this practice demonstrates that LLMs possess sufficient learning and approximation capabilities from data to rival human experience accumulation and pattern learning abilities, at least for handling some common fault types.
Its advantages mainly lie in:
\textbf{(1).} 
Optimize personnel structure to maximize the utilization of senior subsystem experts' expertise, reserving their involvement exclusively for resolving extreme or intractable faults. 
Common fault types can be effectively managed by junior assistants and the SpaceHMchat.
\textbf{(2).} 
Provide interpretability. 
Even senior subsystem experts may require some time-series classification algorithms to aid their judgment, whereas these algorithms fail to offer decision rationale.
Regarding specific implementation techniques, the effectiveness of fine-tuning and LoRA is satisfactory, as shown in Figure \ref{Fig_Results of SpaceHMchat on Fault Localization Task}, where the base model (Qwen3-14B) performs poorly and fails to reach acceptable conclusions in such challenging tasks.
For more precise results, refer to Section \ref{Results of All-in-loop Health Management} or the confusion matrix provided in Figure \ref{Fig_confusion matrix of SpaceHMchat on Fault Localization Task} within \ref{Appendix: Confusion Matrix Result on Fault Localization Task}, which demonstrate that SpaceHMchat achieves high precision in identifying anomaly types.

\subsection{Results of Maintenance Decision-making}
\label{Results of Maintenance Decision-making}

\begin{figure}[htbp]
	\centering
	\begin{minipage}[c]{0.73\textwidth}
		\begin{spacing}{1.2} % 行距倍率
			\begin{chat}

				\chatdate{SpaceHMchat}{}

				\chatpersonOne{
					A fault of "Load 2 open circuit" has been detected in the spacecraft power system. Please execute the root cause analysis and maintenance decision-making task
				}

				\chatpersonTwo{
					\textcolor{red}{\textbf{Step 1 - Root Cause Analysis}}: Reviewing historical analysis reports, archived expert consultations, ..., the possible root causes leading to the "Load 2 open circuit" are summarized as follows:
					\\
					\textbf{1. Connector loosening or aging failure}, i.e., the connector of Load 2 power supply line (e.g., Picoblade connector) may experience material fatigue and increased contact resistance due to in-orbit thermal cycling ..., eventually leading to an open circuit \textcolor{blue}{\textbf{$[5]$}};
					\\
					\textbf{2. High-energy particle impact or space radiation interference}, i.e., cosmic rays or high-energy particles from solar storms may penetrate the spacecraft's shielding, causing circuit components of Load 2 (e.g., MOSFET transistors, relays) to experience single-event latch-up (SEL) or latch-up effects, ..., leading to open circuit failures \textcolor{blue}{\textbf{$[12]$}};
					\\
					\textbf{3. ...};
					\\
					...
					\\
					\textcolor{red}{\textbf{Step 2 - Risk Assessment}}: 
					Reviewing historical fault cases of spacecraft power systems, similar situations include: 
					\\
					\textbf{1. In 2021, the ESA Sentinel-1B spacecraft} experienced ..., resulting in radar data transmission interruption and premature mission termination \textcolor{blue}{\textbf{$[10]$}}; 
					\\
					\textbf{2. In 2005, the ...} experienced ..., leading to ... \textcolor{blue}{\textbf{$[X]$}};
					\\
					...
					\\
					Additionally, analyzing related analysis reports, technical documents, ..., 
					\textcolor{blue}{\textbf{$[7]$}} indicates that similar power failures have led to satellite failures,
					\textcolor{blue}{\textbf{$[16]$}} indicates that ...
					\\
					...
					\\
					\textit{\textbf{In conclusion,} the potential consequences of this type of fault include: Load 2 mission interruption, overall power balance disruption, insufficient power supply to critical payloads.
					Therefore, it is recommended to classify this fault as \textbf{$\llbracket$ \textbf{severe} $\rrbracket$.}}
					\\
					\textcolor{red}{\textbf{Step 3 - Maintenance Strategies}}:
					Reviewing maintenance strategy folders, archived expert consultations, related papers and technical reports, ..., the possible maintenance strategies for this fault are summarized as follows:
					\\
					\textbf{1. Activation of redundant backup units}, i.e., if the design includes redundant power paths for Load 2, activate the redundant branch to ensure continuous operation of Load 2 \textcolor{blue}{\textbf{$[4]$}};
					\\
					\textbf{2. System reboot and parameter reset}, upon detection of the SEL, interrupt the power supply for a duration exceeding 100 ms, followed by a system restart. The repair success rate exceeds 90\% \textcolor{blue}{\textbf{$[27]$}};
					\\
					\textbf{3. ...};
					\\
					...
					\\
					\textcolor{red}{\textbf{Step 4 - References}}:
					\\
					\text{[5]} XXXX Root Cause Analysis Report - xxxx-xx-xx.docx
					\\
					\text{[12]} https://ntrs.nasa.gov/citations/xxxxxxxx
					\\
					...
				}
			\end{chat}
		\end{spacing}
	\end{minipage}
	\hfill
	\begin{minipage}[c]{0.26\textwidth}
		\begin{spacing}{1.2} % 行距倍率
			\begin{chat}

				\chatdate{Base Model}{}

				\chatpersonOne{
					A fault of "Load 2 open circuit" has been detected in the spacecraft power system. Please execute the root cause analysis and maintenance decision-making task
				}

				\chatpersonThree{
					\textbf{Root Cause Analysis and Maintenance Decision-making for "Load 2 Open Circuit"}
					\\
					\textcolor{red}{\textbf{1. Root Cause Analysis}}
					An open circuit in Load 2 indicates a break in the electrical pathway, preventing current flow. Potential causes include:
					\\
					a. Physical Disconnection: ...
					\\
					b. Component Failure: ...
					\\
					...
					\\
					\textcolor{red}{\textbf{2. Maintenance Decision-making}}
					\\
					a. Immediate Actions: ...
					\\
					b. Repair Options: ...
					\\
					c. Long-Term Actions: ...
					\\
					% ...
					% \\
					\textcolor{red}{\textbf{Conclusion}}
					The "Load 2 open circuit" is likely due to a physical disconnection, component failure, or protection device activation. Immediate steps include isolation and diagnostics, followed by repair or replacement. Long-term actions should address root causes to enhance reliability. Coordinate with the mission team to balance urgency with safety and redundancy constraints.
				}
			\end{chat}
		\end{spacing}
	\end{minipage}

	\caption{Results of SpaceHMchat on Maintenance Decision-making Task.
			Our proposed SpaceHMchat's dialogue results are shown on the left, while the right side shows the response of a base model without any techniques mentioned in Figure \ref{Fig_Flow diagram of our method}.
			An example of AIL HM dialogue with complete information is shown in \ref{Appendix: Complete Chat Examples}.
	}
	\label{Fig_Results of SpaceHMchat on Maintenance Decision-making Task}
	\end{figure}

The results of the maintenance decision-making task are shown in Figure \ref{Fig_Results of SpaceHMchat on Maintenance Decision-making Task}.
The success of this practice demonstrates that LLMs possess sufficient knowledge base querying and synthesis capabilities to transform the traditional ``human team analysis'' model into an ``online consultation assistance'' model.
Its advantages mainly lie in:
\textbf{(1).}
Efficient utilization of textual information. 
Traditional research primarily focuses on extracting information from time series telemetry data, with little attention paid to textual sources. 
In fact, spacecraft design documents, technical reports, expert consultation archives, and other texts accumulate rich experiential knowledge. 
The development of LLM technology enables the efficient use of these crucial resources.
Especially crucially, the more specific and detailed the questions asked, the more targeted and professional the responses obtained.
As such specialized information is often hidden within scattered sections of technical reports, it is challenging for traditional manual analysis to achieve comprehensive retrieval and coverage.
\textbf{(2).}
Simplification and acceleration of the process. 
The traditional “human team analysis” model requires on-orbit response teams to search, read, and analyze large volumes of documentation. 
SpaceHMchat, on the other hand, can quickly query and synthesize relevant information during interactive dialogues, significantly simplifying and accelerating this process.
Regarding specific implementation techniques, RAG and MCP are necessary and effective.
Without them, the base model (Qwen3-235B) can only reason based on knowledge acquired during pre-training, resulting in responses that lack specificity, accuracy, and professionalism.

\subsection{Results of All-in-loop Health Management}
\label{Results of All-in-loop Health Management}

\begin{table*}[htbp] %使用table*环境才能适应双栏
	\caption{Performance of SpaceHMchat in all-in-loop health management.}
	\label{Tab_Performance of SpaceHMchat in all-in-loop health management}

	\begin{threeparttable} % 用threeparttable包裹表格和表注才能实现表注和表格一样长而不是超过

		% 以下三句放在tabularx外部，因为不仅控制表内内容的字体大小，标题和表注也要控制
		% 但是不必超过threeparttable达到table*下方那个程度，因为其中\renewcommand\arraystretch{1.4}属于重新定义表格的，会连后续表格也被影响
		\centering
		\renewcommand\arraystretch{1.4} % 调整行高
		\footnotesize

		% \begin{tabularx}{0.81\linewidth}{cc|cc|cc|cc}	
		\begin{tabularx}{0.99\linewidth}{
			>{\centering\arraybackslash}m{0.3cm}
			>{\centering\arraybackslash}m{4cm}
			% >{\centering\arraybackslash}X
			>{\centering\arraybackslash}m{2cm}
			% >{\centering\arraybackslash}X
			>{\centering\arraybackslash}m{9cm}
			% >{\centering\arraybackslash}X |
			% >{\centering\arraybackslash}m{1.5cm}
			}
			% \centering

			\hline
			\hline

			& \textbf{Metrics} & \textbf{Value} & \textbf{Definition} \\

			% \multicolumn{2}{c}{ } & \multicolumn{2}{|c}{Easy Tasks} & \multicolumn{2}{|c}{Medium Tasks} & \multicolumn{2}{|c}{Hard Tasks} \\ 
			% % \hline
			%   &  & Human & SpaceHMchat & Human & SpaceHMchat & Human & SpaceHMchat \\ 

			% \multicolumn{2}{c}{ } & \multicolumn{3}{|c}{Human} & \multicolumn{3}{|c}{SpaceHMchat} \\ 
			% 	% \hline
			% 	&  & Easy Tasks & Medium Tasks & Hard Tasks & Easy Tasks & Medium Tasks & Hard Tasks \\ 

			\hline
			\hline

			\multirow{6}{*}{\adjustbox{angle=90,lap=0pt-0pt,raise=-3.5cm}{Work Condition Recognition}}

			& Time cost of work condition recognition
			& 2.907 s
			& The average time cost of data analysis, indicator calculation, logical reasoning, work condition decision
			\\

			& Accuracy of work condition recognition
			& 50 $/$ 50
			& The ratio of recognition results consistent with true work condition labels
			\\

			& Step-wise correctness rate 
			& 350 $/$ 350
			& The ratio of correctly inferred reasoning steps to all response steps
			\\

			& Logical coherence 
			& 350 $/$ 350
			& The ratio of reasoning steps which are free of contradictions and maintains internal coherence to all response steps
			\\
			
			& Completeness of reasoning 
			& 349 $/$ 350
			& The ratio of samples where all necessary conditions and evidential steps are covered to all response samples
			\\

			& Decision Consistency 
			& 10 $/$ 10
			& The ratio of samples where conclusions remain stable across 5 rounds to all response samples
			\\

			\hline
			\multirow{7}{*}{\adjustbox{angle=90,lap=0pt-0pt,raise=-3cm}{Anomaly Detection}}

			& Time cost of algorithm selection
			& 0.977s
			& The average time cost of searching algorithms and comparing pros/cons
			\\

			& Time cost of algorithm configuration
			& 2.153s
			& The average time cost of parameter setting and implementation setup
			\\

			& Time cost of model training \& testing
			& $\triangle $
			& The average time cost of training, validation, evaluation, anomaly detection
			\\

			& Effectiveness of anomaly detection
			& $\triangle $
			& Measures such as precision, recall, and F1-score
			\\

			& One-shot success rate 
			& 49 $/$ 50
			& The ratio of successfully completing the task on the first attempt
			\\
			
			& Final success rate 
			& 50 $/$ 50
			& The ratio of ultimately completing the task after multiple iterations
			\\

			& Tool Invocation Accuracy 
			& 250 $/$ 251
			& The percentage of requests that strictly conform to the tool calling syntax format
			\\

			\hline
			\multirow{9}{*}{\adjustbox{angle=90,lap=0pt-0pt,raise=-0cm}{Fault Localization}}

			& Time cost of fault localization
			& 48.938 s
			& The average time cost of data analysis and fault localization
			\\

			& Accuracy
			& 89.43\%
			& The proportion of correctly predicted outcomes over the total number of cases
			\\

			& Precision
			& 90.01\%
			& The proportion of true positive predictions among all positive predictions
			\\

			& Recall
			& 89.50\%
			& The proportion of true positive predictions among all actual positive cases
			\\
			
			& F1-score
			& 89.35\%
			& The harmonic mean of precision and recall
			\\

			& MCC
			& 0.8883
			& Matthews correlation coefficient
			\\

			& Kappa
			& 0.8877
			& Cohen's Kappa statistic
			\\

			\hline
			\multirow{9}{*}{\adjustbox{angle=90,lap=0pt-0pt,raise=-4cm}{Maintenance Decision-making}}

			& Time cost of retrieval and root cause analysis
			& 2 min 7.099 s
			& The average time cost of information retrieval and root cause analysis
			\\

			& Time cost of retrieval and risk assessment
			& 2 min 6.981 s
			& The average time cost of information retrieval and risk assessment
			\\

			& Time cost of retrieval and maintenance suggestions
			& 2 min 7.159 s
			& The average time cost of information retrieval and maintenance suggestions
			\\

			& Task completion rate 
			& 50 $/$ 50
			& The ratio of samples where the answer is helpful for guiding operations or decisions to all response samples
			\\

			& Human effort reduction
			& $\bigtriangledown $
			& The extent to which human workload is reduced
			\\

			& Citation accuracy
			& 939 $/$ 965
			& The ratio of citations that actually exist and can find the original document to all citations provided
			\\
			
			& Error severity impact
			& $\triangleleft $
			& If the error message is responded to, the severity of the error message in the task
			\\

			\hline
			\hline

		\end{tabularx}

		\begin{tablenotes}
			\footnotesize
			\item
			The $\triangle $ indicates that this metric depends on the chosen algorithm and is not related to the performance of SpaceHMchat.
			\item 
			The $\bigtriangledown $ indicates that the reduction in human effort depends on the task complexity; for simple tasks, the reduction is not significant, while for complex tasks, the reduction exceeds 50\%.
			\item
			The $\triangleleft $ indicates that the severity of the error message is not serious, mainly due to the loss of citation sources during text generation, leaving only the specific content, which itself does not have serious errors.
		\end{tablenotes}
	
	\end{threeparttable}
\end{table*}

Some metrics are chosen to evaluate the overall performance of SpaceHMchat in AIL HM, as shown in Table \ref{Tab_Performance of SpaceHMchat in all-in-loop health management}.
\begin{enumerate}[1)]
	\item 
	To validate SpaceHMchat's capability of reasoning and chained thinking in the work condition recognition phase, some logical reasoning time and accuracy metrics are designed.
	The \textit{time cost of work condition recognition} of SpaceHMchat is within 3 seconds, indicating that its reasoning is not time-consuming.
	In the random test of 50 AIL HM tasks, there were a total of 350 reasoning steps in the work condition recognition phase, achieving a 100\% \textit{step-wise correctness rate}, \textit{logical coherence}, and \textit{conclusion accuracy}.
	Only one step of the 350 steps outputted additional redundant information, but it did not affect the reasoning logic and conclusion correctness.
	Moreover, in 10 repetitive tests with the same sample, all conclusions were stable and consistent, ensuring \textit{decision consistency}.

	\item
	To validate SpaceHMchat's capability of function calling and tool integration in the anomaly detection phase, some metrics reflecting tool invocation time and success rate are designed.
	The \textit{time cost of algorithm selection} and \textit{time cost of algorithm configuration} of SpaceHMchat are both within 3 seconds, which will improve the work efficiency of operation and maintenance personnel.
	In the random test of 50 AIL HM tasks, it achieved a \textit{one-shot success rate} of 49/50, a \textit{final success rate} of 50/50, and a \textit{tool invocation accuracy} of 250/251.
	It can basically ensure that the entire process of tool calls can be completed smoothly in one attempt, even if there is once failure, only one refresh is needed.

	\item
	To validate SpaceHMchat's capability of learning-and-approximation from data in the fault localization phase, some metrics reflecting localization time and accuracy are designed.
	The \textit{time cost of fault localization} of SpaceHMchat is within 1 minute, mainly consuming time in the extraction of time-frequency features, while the output of conclusion is not time-consuming.
	In the comprehensive test of over 85,000 samples, metrics such as \textit{accuracy}, \textit{precision}, \textit{recall}, and \textit{F1-score} are above 89\%, while \textit{MCC} and \textit{Kappa} are above 0.88 (also visualized in the confusion matrix in Figure \ref{Fig_confusion matrix of SpaceHMchat on Fault Localization Task} in \ref{Appendix: Confusion Matrix Result on Fault Localization Task}).
	Its performance is satisfactory, proving that it is sufficient to handle most common fault localization tasks.

	\item
	To validate SpaceHMchat's capability of knowledge base querying and synthesis in the maintenance decision-making phase, some metrics reflecting information retrieval time, task completion rate, and human effort reduction are designed.
	The \textit{time cost of retrieval and root cause analysis}, \textit{time cost of retrieval and risk assessment}, and \textit{time cost of retrieval and maintenance suggestions} of SpaceHMchat are all within 3 minutes, mainly consuming time in information retrieval.
	This time consumption is acceptable, as even with manual retrieval and information analysis by human experts, the time consumption would exceed 30 minutes, and SpaceHMchat can significantly reduce this workload.
	In the random test of 50 AIL HM tasks, all answers provided in the maintenance decision-making phase can output helpful information to guide operations or decisions, providing assistance to the human team.
	It was confirmed that among the total of 965 references provided in these tests, 939 are explicit and can be directly accessed. Only a small portion cannot be opened directly due to the loss of reference sources during generation. However, these contents do exist in the knowledge base, and no errors or hallucinated generations are involved.
\end{enumerate}

\section{Discussion}
\label{Discussion}

It must be acknowledged that this work remains a forward-looking exploratory study. 
Although it demonstrates the potential and necessity of LLM technologies in the SMC era, numerous limitations still exist.
Particularly in light of the unique attributes such as high reliability requirements, non-transferable responsibility, and absolute human control authority in the aerospace domain, this work emphasizes HAIC rather than the fully automated agents.
SpaceHMchat, guided by the AUC principle (Figure \ref{Fig_SpaceHMchat: a Human-AI collaboration framework for AIL HM, guided by the AUC principle}), can currently only be positioned as a copilot or sub-task agent.
In the future, assuming that the capabilities of LLMs exceed human and that issues of responsibility attribution are resolved, the principle of surpassing human underlying capabilities (SUC principle) rather than AUC principle will lead to reliable fully automated agents:
\text{(1).}
In the work condition recognition phase, LLMs will no longer rely on decision trees but will build an interactive digital twin based on design documents, real-time reasoning the dynamic changes in work status, automating task scheduling and attitude trajectory control.
\text{(2).}
In the anomaly detection phase, an agent will autonomously maintain one or a group of spacecraft, capable of trying various algorithms, improving and training models, selecting the optimal algorithm for monitoring, and generating detection reports.
\text{(3).}
In the fault localization phase, LLMs will update the fault causal knowledge graph in real-time based on fault case data and other information, not only locating the fault type but also providing the fault's propagation path, development trend, and causal link visualization.
\text{(4).}
In the maintenance decision-making phase, humans will grant LLMs the authority to access the spacecraft information system, automatically maintaining the knowledge base and directly driving telemetry operation commands for repairing and event archiving.

\section{Conclusion}
\label{Conclusion}

This work attempts to explore:
\textit{Can we develop a HAIC framework, empowered by LLMs, to adapt the AIL HM of SPS in the SMC era, unlocking solutions to unprecedented challenges?}
Based on the developed SpaceHMchat and the AIL experimental validation system, we demonstrate the potential and necessity of HAIC in the future SMC era.
The experimental results demonstrate that SpaceHMchat achieves excellent performance across 23 quantitative indicators, including 100\% conclusion accuracy in logical reasoning of work condition recognition, over 99\% success rate in anomaly detection tool invocation, over 90\% precision in fault localization, and maintenance decision knowledge base search within 3 minutes.
Through conversational interaction, SpaceHMchat links maintenance personnel, time-series analysis programs, telemetry data, and aerospace documentations, freeing operators from intricate technical details and allowing them to focus on high-level decision-making and supervision.
The future direction is to evolve to a stable, reliable, fully automated agent, exploring the SUC principle.
Additionally, this work calls for more emphasis on the accumulation and management of aerospace textual documentation, while previous studies have focused on developing advanced analysis algorithms for sensor signals.

\section*{Resource availability}
% Data and code availability:
All data have been deposited at Hugging Face and are publicly available as of the date of publication at \url{https://huggingface.co/datasets/DiYi1999/XJTU-SPS}.
All original code has been deposited at Github and is publicly available at \url{https://github.com/DiYi1999/SpaceHMchat} as of the date of publication.
Meanwhile, the detailed introduction information and download links of the datasets are available at \url{https://diyi1999.github.io/XJTU-SPS/}.

% All data have been deposited at Mendeley Data and are publicly available as of the date of publication at \url{https://doi.org/10.17632/6nv524tczf.1}.
% All original code has been deposited at Mendeley Data and is publicly available at \url{https://doi.org/10.17632/6gjy6j8vpm.1} as of the date of publication.
% Meanwhile, the detailed introduction information and download links of the datasets and codes are also publicly available as of the date of publication at \url{https://diyi1999.github.io/XJTU-SPS/} and \url{https://github.com/DiYi1999/SpaceHMchat}.

\section*{Acknowledgment}
This work is supported in part by the National Natural Science Foundation of China under Grant 52475128 and Grant U22B2014.

\section*{Author contributions}
Writing - original draft: Y. D.;
Writing - review \& editing: Z. Zhao, F. W., X. L., J. T., J. R. and Z. Zhai;
Conceptualization: Y. D. and Z. Zhao;
Methodology: Y. D.;
Validation: Y. D.;
Visualization: Y. D.;
Data curation: Y. D.;
Supervision: Z. Zhao and Z. Zhai;
Resources: Z. Zhao and Z. Zhai;
Funding acquisition: Z. Zhao, Z. Zhai, and X. C.;
Project administration: X. C.

\section*{Declaration of interests}
The authors declare no competing interests.

% 再添加一个附录
\section*{Supplemental information}
% \section*{Appendix}
The appendix is available on the journal's supplementary material page or at \url{https://github.com/DiYi1999/SpaceHMchat}.

\bibliographystyle{elsarticle-num}
\bibliography{Reference}

%\includepdf[pages={1-5}, angle=0]{Tables.pdf}

%\includepdf[pages={1-21}, angle=90]{Results.pdf}

% filepath: appendix.tex
\appendix

\section{Complete Chat Examples}
\label{Appendix: Complete Chat Examples}

The complete dialogue example of SpaceHMchat is shown on the next page, which is directly captured from the dialogue software screenshot.
Currently, SpaceHMchat mainly based on Chinese conversations. 
To facilitate reading, we have attached English translations after each round of Chinese questions and answers.

About the reason why the current SpaceHMchat mainly uses Chinese for conversations, there are three main considerations:
(1) This project is supported by a Chinese fund project and is currently dedicated to serving the user group in the Chinese aerospace field.
(2) We must ensure that the base LLM models used are open source, which can be freely downloaded and deployed by users while ensuring information security. Therefore, we chose the Qwen series models (Qwen-14B, Qwen-235B) as our base models, as Qwen models are among the few open-source large models that can rival closed-source models such as ChatGPT. 
Moreover, Qwen models are trained on a large amount of Chinese corpus, thus performing better in Chinese conversations.
(3) Our knowledge base mainly consists of Chinese materials, including a large number of Chinese technical documents, papers, and reports. Therefore, it can better utilize these knowledge base contents in Chinese conversations.

% 把长图作为独立一页插进来，宽度自动铺满纸张

% \includepdf[
%     pages=-,            % 全部页
%     fitpaper=true,      % 让这一页的尺寸=纸张尺寸
%     noautoscale=false   % 允许 pdfpages 自动缩放
% ]{CompleteChatExample-cut8pages.pdf}



你好，我是默认助手。你可以立刻开始和我聊天

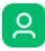
**DiYi1999**
01/08 21:21

你可以为我做什么?
(What can you do for me?)

Tokens: 17

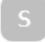
**SpaceHMchat | dy@ssh62229**
01/08 21:48

> 已深度思考（用时 3.0 秒） ›

我能够为您执行以下五类任务，请以对应格式告知需要执行的任务类型:

【请以以下格式明确告知您需要执行的任务：请执行<普通问答/异常检测/工况识别/故障定位/根因分析及维护决策>任务...】

1. **普通问答**：回答航天器电源系统的原理、设计、运维知识问题（例如电池寿命、配电策略、故障代码含义等）。
2. **异常检测**：提供算法选择（如孤立森林、LSTM）、参数调优、数据训练与测试，生成异常评分曲线及报警阈值建议。
3. **工况识别**：基于传感器数据（如电压、电流、温度）分析当前设备状态（如放电模式、太阳能板展开状态等）。
4. **故障定位**：调用微调模型判别故障类型（如短路、开路、老化）并定位部件模块（如电池模块、DC-DC变换器）。
5. **根因分析及维护决策**：结合知识库分析故障根本原因（如焊接缺陷、辐射损伤），输出风险等级、隔离方案、在轨/地面修复建议及备件推荐。

请直接指定任务类型及对应输入（如问题描述/数据段/知识库路径），我将一次性输出分析结果与解决方案。

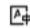

I can perform the following five types of tasks for you; please use the corresponding format to indicate the task type needed:

【Please clearly specify the task you need in the following format: Please perform the <General Q&A/Anomaly Detection/Operating State Recognition/Fault Localization/Root Cause Analysis and Maintenance Decision-Making> task...】

1. **General Q&A**: Answer questions about the principles, design, and operations and maintenance of spacecraft power systems (e.g., battery life, power distribution strategies, meanings of fault codes, etc.).
2. **Anomaly Detection**: Provide algorithm selection (e.g., Isolation Forest, LSTM), parameter tuning, data training and testing, and generate anomaly score curves and alarm threshold recommendations.
3. **Operating State Recognition**: Based on sensor data (e.g., voltage, current, temperature), analyze the current device status (e.g., discharge mode, solar array deployment status, etc.).
4. **Fault Localization**: Invoke fine-tuned models to determine fault types (e.g., short circuit, open circuit, aging) and locate the component level (e.g., battery module, DC-DC converter).
5. **Root Cause Analysis and Maintenance Decision-Making**: Leverage the knowledge base to analyze the root causes of faults (e.g., welding defects, radiation damage), and output risk level, isolation plan, on-orbit/ground repair recommendations, and spare parts recommendations.

Please directly specify the task type and the corresponding input (e.g., problem description/data segment/knowledge base path), and I will output the analysis results and solutions in a single response.

Tokens: 722 · 117 ↑703

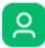
**DiYi1999**
01/09 18:34

世界各国的巨型卫星星座计划都有哪些，各自计划部署多少航天器? 用表格之类的总结一下。
(What are the mega satellite constellation plans in the world, and how many spacecraft do they plan to deploy respectively?)

Tokens: 69

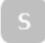
**SpaceHMchat | dy@ssh62229**
01/09 18:35

> 已深度思考（用时 8.6 秒） ›

请执行<普通问答>任务:

以下是截至2024年的全球主要巨型卫星星座计划总结（基于公开文献及行业报告）:

| 星座名称 | 国家/公司 | 计划部署数量 | 当前部署数量 | 轨道高度（km） | 频段 | 状态 | 用途 |
|---|---|---|---|---|---|---|---|
| Starlink | SpaceX | 12,000 | ~6,000 (2024) | 550（LEO） | Ku/Ka/Band | 部署中 | 宽带互联网 |
| Project Kuiper | Amazon | 3,236 | 2（原型星） | 610-630（LEO） | Ka | 计划发射 | 商用低延迟通信 |
| OneWeb | 英国Bharti集团 | 6,372 | ~600 (2024) | 1,200（LEO） | Ku | 暂停恢复后 | 全球互联网覆盖 |
| 中国"银河" | 中国航科+中国星网 | 10,000 | 0（规划阶段） | 500-1,200（LEO） | 多频段 | 研制中 | 5G通信、导航增强 |
| Telesat LEO | 加拿大Telesat | 1,671 | 2（原型星） | 1,000（LEO） | Ka | 工程验证阶段 | 企业级高速通信 |
| Sphere | 俄罗斯Roscosmos | 640 | 0（规划阶段） | 12,000（MEO） | 未知 | 计划审议 | 军民两用通信 |
| 韩国Kuiper | 韩国KAI+三星 | 2,000 | 未披露 | LEO/MEO | 未知 | 研究阶段 | 6G通信技术储备 |
| 印度BharatNet | 印度Devas公司 | 200+ | 0 | 600（LEO） | Ka/Ku | 审批中 | 农村宽带覆盖 |

**补充说明:**

1. **监管争议**：各国计划中约30%面临国际电信联盟（ITU）频谱/轨道资源分配冲突，尤其是LEO轨道密度限制争议;
2. **技术挑战**：SpaceX与OneWeb早期计划已引发空间碎片治理争议（NASA预警2025年太空交通管制压力增大）;
3. **电源系统关联**：此类巨型星座对航天器电源系统的寿命（>5年）、抗辐射能力、自主管理需求提出新标准（如GaN器件应用、智能功率分配）。

如需获取具体卫星电源设计案例（如Starlink的钙化氮太阳能板效率、OneWeb的锂离子电池热控方案），请补充数据后说明。

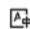

**Please perform the <General Q&A> task:**

Below is a summary of the world's major mega-satellite constellation plans as of 2024 (based on public literature and industry reports):

| Constellation Name | Country/Company | Planned Number of Satellites | Currently Deployed | Orbital Altitude (km) | Frequency Band | Status | Use |
|---|---|---|---|---|---|---|---|
| Starlink | SpaceX | 12,000 | ~6,000 (2024) | 550 (LEO) | Ku/Ka/Band | Being deployed | Broadband internet |
| Project Kuiper | Amazon | 3,236 | 2 (prototype) | 610-630 (LEO) | Ka | Planned launch | Commercial low-latency communications |
| OneWeb | UK Bharti Group | 6,372 | ~600 (2024) | 1,200 (LEO) | Ku | Restarted after suspension | Global internet coverage |
| China "Galaxy" | China Aerospace Science and Technology + China Star Net | 10,000 | 0 (planning stage) | 500-1,200 (LEO) | Multi-band | In development | 5G communications, navigation augmentation |
| Telesat LEO | Canada Telesat | 1,671 | 2 (prototype) | 1,000 (LEO) | Ka | Engineering validation phase | Enterprise-grade high-speed communications |

| Sphere | Russia Roscosmos | 640 | 0 (planning stage) | 12,000 (MEO) | Unknown | Under review | Dual-use military and civilian communications |
| Korea Kuiper | Korea KAI + Samsung | 2,000 | Not disclosed | LEO/MEO | Unknown | Research phase | 6G communication technology reserve |
| India BharatNet | India Devas Company | 200+ | 0 | 600 (LEO) | Ka/Ku | Under approval | Rural broadband coverage |

Supplementary notes:

1. Regulatory disputes: about 30% of the plans face International Telecommunication Union (ITU) spectrum/orbital resource allocation conflicts, especially disputes over LEO orbital density limits;

2. Technical challenges: early plans by SpaceX and OneWeb have already raised space debris governance controversies (NASA warns of a sharp increase in space traffic management pressure in 2025);

3. Power system implications: such mega-constellations impose new standards on spacecraft power system lifetime (>5 years), radiation tolerance, and autonomy requirements (e.g., GaN device applications, intelligent power distribution).

If you need specific satellite power design cases (e.g., Starlink's gallium arsenide solar panel efficiency, OneWeb's lithium-ion battery thermal control scheme), please provide additional data access authorization.

Tokens: 2349 ↑791 ↓1358

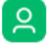 **DiYi1999**
01/08 22:30

请执行工况识别任务，在 /data/DiYi/DATA/Our_Exp_Data/XJTU-SPS Dataset/XJTU-SPS for MR/XJTU-SPS for MR_94cycles.csv 这个文件中，检查 2024/10/22 19:38:27 周围几个点，判断航天器正处于什么工况？
(Please perform the work condition recognition task in the file /data/DiYi/DATA/Our_Exp_Data/XJTU-SPS Dataset/XJTU-SPS for MR/XJTU-SPS for MR_94cycles.csv. Check a few timestamps around 2024/10/22 19:38:27 to determine what condition the spacecraft is in?)

Tokens: 194

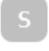 **SpaceHMchat | dy@ssh62229**
01/08 22:43

> 💡 已深度思考（用时 20.3 秒）

步骤1：确认太阳能电池板电压是否大于1V？太阳能电池板电压数据（20.914,20.871,20.937）均远超1V，答案为是，记录为日照区，转入步骤2；

步骤2：确认各路负载电流是否大于0.5A？负载1电流数据（2.544,2.544,2.544）均大于0.5A，负载2-3电流均为0或接近0，答案为是，记录为正在执行任务1，转入步骤3；

步骤3：确认负载1功率是否大于太阳能电池板功率？负载1功率数据（29.134,29.093,29.078）均大于太阳能电池板功率数据（17.7,17.8,17.7），答案为是，输出工作状态为【联合供电、日照区、任务1】，转入步骤7；

步骤7：根据所输出工作状态检查各传感器观测值：太阳能电池板参数（电压20.8-21.0V<35V、电流0.845-0.851A<2.7A、功率17.7-17.8W<70W）；负载总参数（电压10.95-10.97V<13V、总功率26.287-26.338W<27W）；BCR参数（电压16.415-16.513V、电流1.003A<3.5A、功率16.563W<58W）；电池组2-4参数（电压16.401-16.488V、电流0.144-0.423A在-0.7~+0.5A内、温度26.4-26.6℃）；母线参数（电压16.213-16.303V、电流1.851A<2A）；负载1参数（电压11.43-11.452V、电流2.544A<3A、温度37.5℃）；负载2-3参数（电流均为0或0.003A、功率0.016W<16W），检查正常；

综上，该航天器目前的工作状态为：【联合供电、日照区、任务1】。

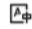

Step 1: Confirm whether the solar panel voltage is greater than 1V? Solar panel voltage data (20.914,20.871,20.937) all far exceed 1V, answer is yes, recorded as sunlit region, proceed to Step 2;

Step 2: Confirm whether each load current is greater than 0.5A? Load 1 current data (2.544,2.544,2.544) are all greater than 0.5A, loads 2-3 currents are all 0 or near 0, answer is yes, recorded as performing Task 1, proceed to Step 3;

Step 3: Confirm Load 1 power is greater than the solar panel power? Load 1 power data (29.134,29.093,29.078) are all greater than the solar panel power data (17.7,17.8,17.7), answer is yes, output operating state as [Combined Power Supply, Sunlit Region, Task 1], proceed to Step 7;

Step 7: Check sensor observations according to the output operating state: Solar panel parameters (voltage 20.8–21.0V<35V, current 0.845–0.851A<2.7A, power 17.7–17.8W<70W); load total parameters (voltage 10.95–10.97V<13V, total power 26.287–26.338W<27W); BCR parameters (voltage 16.415–16.513V, current 1.003A<3.5A, power 16.563W<58W); battery packs 2–4 parameters (voltage 16.401–16.488V, current 0.144–0.423A within -0.7~+0.5A, temperature 26.4–26.6℃); bus parameters (voltage 16.213–16.303V, current 1.851A<2A); Load 1 parameters (voltage 11.43–11.452V, current 2.544A<3A, temperature 37.5℃); Loads 2–3 parameters (currents both 0 or 0.003A, power 0.016W<16W), checks normal;

In summary, the spacecraft's current operating state is: [Combined Power Supply, Sunlit Region, Task 1].

Tokens: 4269 ↑2311 ↓1958

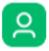 **DiYi1999**
01/08 22:44

请执行异常检测任务，你有什么算法可供我选择？
(Please execute the anomaly detection task. What algorithms do you have for me to choose from?)

Tokens: 45

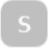 **SpaceHMchat | dy@ssh62229**
01/08 23:31

检查了本地的算法库，可供用于进行异常检测的算法列表为：['Autoformer', 'DLinear', 'FourierGNN', 'GAT', 'GCN', 'GIN', 'GRU', 'GraphWaveNet', 'Informer', 'MLP', 'MTGNN', 'Nothing', 'PatchTST', 'SGC', 'StemGNN', 'TCN', 'Transformer']。

以下是一些关于这些算法的参考信息，可以帮助您进行选择：

1. **适用场景**：
   - 如果您的数据是时间序列且依赖性强，考虑使用 `Informer` 或 `Autoformer`；
   - 针对图结构数据，可以优先选择 `GAT`、`GCN`、`SGC` 和 `GraphWaveNet`；
   - 如果需要简单的模型实现快速实验，`MLP` 和 `DLinear` 非常适合；
   - 非图结构的时间序列数据适合采用 `GRU`、`TCN` 和 `StemGNN`，前者侧重序列建模，后者可以有效捕捉时间序列中的谱特性；
   - `MTGNN` 是图与时间序列结合的一种算法，适用于多变量时间序列预测问题。

2. **优劣势分析**：
   - **Transformer 系列**（如 Informer、Autoformer）：适合长序列建模，但可能对计算资源要求较高。
   - **图神经网络**（如 GAT、GCN）：在处理具有复杂依赖关系的异构图数据时表现出色，但可能需要多调整参数的时间。
   - **轻量级算法**（如 MLP、DLinear）：速度快，实现简单，但可能在复杂模型建模上表现有所限。

3. **推荐建议**：
   - 如果您的任务是基于图结构的时间序列分析（如传感器网络监控数据），可以选择 `MTGNN` 或者 `GraphWaveNet`。
   - 对于单纯的时间序列异常检测任务，可以尝试使用 `Informer` 或 `StemGNN`，两者分别通过高效的注意力机制和谱建模对时间序列进行高效建模。
   - 如果对模型复杂度有严格要求，可以优先考虑使用 `DLinear` 或简化模型 `MLP` 做初步验证。

根据任务性质、数据结构和资源条件，您可以选择最适合的算法来开始工作。如果需要更多信息或帮助选择特定算法的参数配置，请随时告诉我！

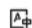

Checked the local algorithm library, the list of algorithms available for anomaly detection is: ['Autoformer', 'DLinear', 'FourierGNN', 'GAT', 'GCN', 'GIN', 'GRU', 'GraphWaveNet', 'Informer', 'MLP', 'MTGNN', 'Nothing', 'PatchTST', 'SGC', 'StemGNN', 'TCN', 'Transformer'].

Below is some reference information about these algorithms, which can help you make a choice:

1. **Applicable Scenarios:**
   - If your data is time series and has strong dependencies, consider using `Informer` or `Autoformer`;
   - For graph-structured data, prioritize `GAT`, `GCN`, `SGC`, and `GraphWaveNet`;
   - If a simple model is needed for quick experiments, `MLP` and `DLinear` are very suitable;
   - Non-graph-structured time series data is suitable for `GRU`, `TCN`, and `StemGNN`. The former focuses on sequence recursive modeling, while the latter can effectively capture spectral characteristics in time series;
   - `MTGNN` is an algorithm combining graph and time series, suitable for multi-variable time series forecasting problems.

2. **Advantages and Disadvantages Analysis:**
   - **Transformer Series** (e.g., Informer, Autoformer): Suitable for long sequence modeling, but may require high computational resources.
   - **Graph Neural Networks** (e.g., GAT, GCN): Perform well in handling heterogeneous graph data with complex dependencies but require more time for parameter tuning.
   - **Lightweight Algorithms** (e.g., MLP, DLinear): Fast, simple to implement, but may be limited in modeling complex patterns.

3. **Recommendation Suggestions:**
   - If your task involves graph-structured time series analysis (e.g., sensor network monitoring data), you can choose `MTGNN` or `GraphWaveNet`.
   - For pure time series anomaly detection tasks, try using `Informer` or `StemGNN`. The former models time series efficiently through effective attention mechanisms, while the latter does so through spectral modeling.
   - If there are strict requirements on model complexity, consider using `DLinear` or the simplified model `MLP` for preliminary validation.

Based on the nature of the task, data structure, and resource conditions, you can select the algorithm that best fits your needs to start working. If you need more information or assistance in selecting specific algorithm parameter configurations, please let me know!

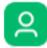

**DiYi1999**
01/08 22:51

我希望选用MTGNN，有什么需要我设置的参数吗？
(I hope to choose MTGNN. Are there any parameters that I need to set?)

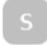

**SpaceHMchat | dy@ssh62229**
01/08 23:33

指定的异常检测算法 MTGNN 需要设置的参数为：

1. `AD_threshold`：异常分数阈值，误差超过这个值就认为是异常 (默认值: 0.5)。
2. `Decompose`：预处理多尺度分解方法 (默认值: None)。
3. `LeakyReLU_slope`：LeakyReLU斜率 (默认值: 0.2)。
4. `Method`：项目名称 (默认值: SPS_AD_LLM_Project)。
5. `TASK`：任务类型 (默认值: anomaly_detection)。
6. `Version`：实验编号 (默认值: V1.0)。
7. `Wavelet_level`：小波分解层数 (默认值: 2)。
8. `batch_size`：批处理大小 (默认值: 1024)。
9. `block_residual`：是否对空间模块使用残差连接 (默认值: False)。
10. `data_name`：数据集名称 (默认值: XJTU-SPS for AD)。
11. `dataset_tra_d_val`：训练集与验证集划分比例 (默认值: 8)。
12. `devices`：使用的设备，GPU编号列表 (默认值: [0])。
13. `dropout`：Dropout率 (默认值: 0.05)。
14. `gradient_clip_val`：梯度裁剪，防止梯度爆炸 (默认值: 100.0)。
15. `if_add_work_condition`：是否使用工况信息也作为输入信息的一部分 (默认值: False)。
16. `if_timestamp`：是否使用时间戳也作为输入信息的一部分 (默认值: True)。
17. `label_len`：Transformers目回目起始标签长度 (默认值: 1)。
18. `lag`：序列长度 (默认值: 96)。
19. `lag_step`：滑动窗口步长 (默认值: 1)。
20. `lr`：学习率，如0.0001 (默认值: 0.0001)。
21. `patience`：连续多少个epoch内loss都没有减下去就停止训练 (默认值: 5)。
22. `pred_len`：预测长度 (默认值: 1)。
23. `reco_form`：此参数已弃用 (默认值: all_to_all)。
24. `result_root_path`：结果保存路径 (默认值: /data/DiYi/MyWorks_Results)。
25. `scale`：是否执行标准化预处理 (默认值: True)。
26. `scheduler`：使用的学习率衰减策略，可选项有ReduceLROnPlateau、StepLR、ExponentialLR、CosineAnnealingLR、CosineAnnealingWarmRestarts (默认值: ReduceLROnPlateau)。
27. `spatial_block`：空间信息捕捉模块，目前可选项有GCN、GAT、GIN、SGC、MTGNN、FourierGNN、StemGNN、GraphWaveNet、Nothing (默认值: MTGNN)。
28. `temporal_block`：时间信息捕捉模块，目前可选项有TCN、GRU、Transformer、Informer、Autoformer、PatchTST、DLinear、Nothing (默认值: Nothing)。
29. `MTGNN_buildA_true`：是否构建自适应的A矩阵 (默认值: True)。
30. `MTGNN_conv_channels`：TC模块输入维度是residual_channels，输出维度是conv_channels (默认值: 32)。
31. `MTGNN_dilation_exponential`：膨胀指数 (默认值: 1)。
32. `MTGNN_end_channels`：end_conv_1的输出维度 (默认值: 128)。
33. `MTGNN_gcn_depth`：mixprop里面的图卷积深度 (默认值: 2)。
34. `MTGNN_gcn_true`：是否增加GCN层 (默认值: True)。
35. `MTGNN_graph_k`：每个节点几个邻居节点 (默认值: 24)。
36. `MTGNN_in_dim`：输入维度，设置为1即可 (默认值: 1)。
37. `MTGNN_layer_norm_affline`：在层归一化中是否进行逐元素仿射操作 (默认值: True)。
38. `MTGNN_layers`：多少层GC + TC (默认值: 3)。
39. `MTGNN_node_embedding`：节点嵌入的维度 (默认值: 192)。
40. `MTGNN_propalpha`：Prop alpha，即在混合传播中保留根节点原始状态的比例，取值范围在0到1之间 (默认值: 0.05)。
41. `MTGNN_residual_channels`：GC模块输入维度是conv_channels，输出维度是residual_channels (默认值: 32)。
42. `MTGNN_skip_channels`：跳跃连接的输出维度。（输入维度是conv_channels）(默认值: 64)。
43. `MTGNN_tanhalpha`：生成邻接矩阵时的的双曲正切alpha值，alpha控制饱和度和率 (默认值: 3)。

## 参数设置建议

1. 关键参数：

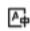

The parameters that need to be set for the specified anomaly detection algorithm MTGNN are:

1. $AD\_threshold$ : Anomaly score threshold, errors exceeding this value are considered anomalies (default: 0.5).
2. $Decompose$ : Multiscale decomposition method for preprocessing (default: None).
3. $LeakyReLU\_slope$ : LeakyReLU slope (default: 0.2).
4. $Method$ : Project name (default: SPS_AD_LLM_Project).
5. $TASK$ : Task type (default: anomaly_detection).
6. $Version$ : Experiment number (default: V1.0).
7. $Wavelet\_level$ : Wavelet decomposition levels (default: 2).
8. $batch\_size$ : Batch size (default: 1024).
9. $block\_residual$ : Whether to use residual connections for spatial modules (default: False).
10. $data\_name$ : Dataset name (default: XJTU-SPS for AD).
11. $dataset\_tra\_d\_val$ : Training and validation set split ratio (default: 8).
12. $devices$ : Devices used, list of GPU IDs (default: [0]).
13. $dropout$ : Dropout rate (default: 0.05).
14. $gradient\_clip\_val$ : Gradient clipping to prevent gradient explosion (default: 100.0).
15. $if\_add\_work\_condition$ : Whether to include operating conditions as part of the input information (default: False).
16. $if\_timestamp$ : Whether to include timestamps as part of the input information (default: True).
17. $label\_len$ : Start label length for Transformers' autoregression (default: 1).
18. $lag$ : Sequence length (default: 96).
19. $lag\_step$ : Sliding window step size (default: 1).
20. $lr$ : Learning rate, default 0.0001 (default: 0.0001).
21. $patience$ : Stop training if loss does not decrease after a certain number of consecutive epochs (default: 5).
22. $pred\_len$ : Prediction length (default: 1).
23. $reco\_form$ : This parameter is deprecated (default: all_to_all).
24. $result\_root\_path$ : Result saving path (default: /data/DiYi/MyWorks_Results).
25. $scale$ : Whether to perform standardization preprocessing (default: True).
26. $scheduler$ : Learning rate decay strategy used, options include ReduceLROnPlateau, StepLR, ExponentialLR, CosineAnnealingLR, CosineAnnealingWarmRestarts (default: ReduceLROnPlateau).
27. $spatial\_block$ : Spatial information capture module, currently options include GCN, GAT, GIN, SGC, MTGNN, FourierGNN, StemGNN, GraphWaveNet, Nothing (default: MTGNN).
28. $temporal\_block$ : Temporal information capture module, currently options include TCN, GRU, Transformer, Informer, Autoformer, PatchTST, DLinear, Nothing (default: Nothing).
29. $MTGNN\_buildA\_true$ : Whether to build an adaptive A matrix (default: True).
30. $MTGNN\_conv\_channels$ : Input dimension of the TC module is residual_channels, output dimension is conv_channels (default: 32).
31. $MTGNN\_dilation\_exponential$ : Dilation exponential (default: 1).
32. $MTGNN\_end\_channels$ : Output dimension of end_conv_1 (default: 128).
33. $MTGNN\_gcn\_depth$ : Graph convolution depth in mixprop (default: 2).
34. $MTGNN\_gcn\_true$ : Whether to add GCN layers (default: True).
35. $MTGNN\_graph\_k$ : Number of neighboring nodes per node (default: 24).
36. $MTGNN\_in\_dim$ : Input dimension, set to 1 (default: 1).
37. $MTGNN\_layer\_norm\_affline$ : Whether to perform element-wise affine operations in layer normalization (default: True).
38. $MTGNN\_layers$ : Number of GC + TC layers (default: 3).
39. $MTGNN\_node\_embedding$ : Dimension of node embedding (default: 192).
40. $MTGNN\_propalpha$ : Prop alpha, the proportion of retaining the root node's original state in mixed-hop propagation, with a range between 0 and 1 (default: 0.05).
41. $MTGNN\_residual\_channels$ : Input dimension of the GC module is conv_channels, output dimension is residual_channels (default: 32).
42. $MTGNN\_skip\_channels$ : Output dimension of skip connections (input dimension is conv_channels) (default: 64).
43. $MTGNN\_tanhalpha$ : Hyperbolic tangent alpha value for generating the adjacency matrix, alpha controls saturation rate (default: 3).

**Parameter Setting Recommendations**

1. **Key Parameters**:

- $AD\_threshold$ (anomaly score threshold) can be manually adjusted based on the historical anomaly distribution of the data.
- $MTGNN\_build2a\_true$ : It is recommended to keep this as $True$ to automatically construct adjacency matrices to adapt to complex graph patterns in the data.
- $MTGNN\_gcn\_true$ : Keep as $True$ , as adding graph convolution layers is very helpful for improving model expression capabilities.
- $MTGNN\_graph\_k$ : If the node association in the graph is not obvious, this value can be reduced appropriately, e.g., reduced to 1/4–1/2 of the total number of nodes.

2. **Training-Related Parameter Adjustments**:
- $batch\_size$ : The default is 1024, but if memory is limited, try reducing it to 64 or 128.
- $lr$ : The default of 0.0001 is a reasonable starting point for most scenarios; if the model does not converge, it can be lowered to 0.00005.
- $patience$ : Controls the early stopping mechanism of the model; if training loss fluctuates greatly, it can be increased to 10 or higher.

3. **Data and Preprocessing Parameters:**
- $scale$ : It is recommended to keep it as $True$ , unless the data is already explicitly normalized or standardized.
- $lag$ : Sequence length can be adjusted based on actual time characteristics (e.g., if the cycle is hourly or daily, try adjusting to 48 or 72).
- $lag\_step$ : Adjust this value based on data density; for dense data, it can be increased to speed up processing, and for sparse data, keep it as 1.

4. **Spatial Module ( $spatial\_block$ ) and Other Component Recommendations:**
- Currently, $MTGNN$ is the default $spatial\_block$ , and it is not recommended to replace it with other modules unless you want to test comparative experiments with different components.
- $temporal\_block$ : In certain tasks, introducing temporal modeling modules such as TCN or Transformer can further improve model performance.
  For example, you can set $temporal\_block: TCN$ to capture long- and short-term temporal features.

5. **Wavelet and Multiscale Decomposition:**
- If the data exhibits multiscale patterns (e.g., high noise, seasonal and periodic changes), wavelet decomposition can be enabled, and $wavelet\_level$ can be increased to 3 or higher to enhance multiscale modeling.

If you need further optimization of MTGNN's parameters, you can also consider phased parameter tuning:

- First, run training and testing with the default settings to observe baseline performance.
- Then, adjust key parameters such as learning rate, adjacency matrix configuration ( $MTGNN\_tanhalpha$ , $MTGNN\_graph\_k$ ), or wavelet levels ( $wavelet\_level$ ).

If you need help with these setups, such as adjusting default parameters or specifying file paths for training and detection tasks, please feel free to let me know!

Tokens: 7997 ↑5150 ↓2047

---

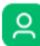 **DiYi1999**
01/08 23:33

把MTGNN算法的batch_size这个参数修改成2048。
(Modify the parameter "batch_size" of the MTGNN algorithm to 2048.)

Tokens: 56

---

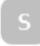 **SpaceHMchat | dy@ssh62229**
01/08 23:33

1. 指定的异常检测算法 'MTGNN' 的参数已经成功设置或修改。

2. 补充信息:
- 参数 $batch\_size$ 已更新为 2048。这意味着在每次训练迭代中,模型将处理更多的样本,可能带来更快的训练速度,但也需要更多的计算资源和显存支持。
- 如果在后续训练中遇到显存不足的情况,建议降低 $batch\_size$ 的值(比如10到1024或更低),以适应硬件资源限制。
- 大 $batch\_size$ 有时可能导致模型收敛速度变慢或泛化性能下降,建议您在训练过程中密切观察训练过程中的损失变化和模型效果,必要时进行调整。

如果您计划进行模型训练或测试,请告知我数据路径,我可以帮助您运行训练(或测试)流程。

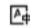

1. The parameters of the specified anomaly detection algorithm 'MTGNN' have been successfully set or modified.

2. Additional information:
- The parameter $batch\_size$ has been updated to 2048. This means that in each training iteration, the model will process more samples, which may result in faster training but will also require more computational resources and GPU memory support.
- If you encounter insufficient GPU memory during subsequent training, it is recommended to lower the value of $batch\_size$ (e.g., to 1024 or lower) to adapt to hardware resource limitations.
- A large $batch\_size$ may sometimes slow down the model's convergence speed or degrade generalization performance. Therefore, you can closely monitor the loss changes and model performance during training and make adjustments if necessary.

If you plan to conduct model training or testing, please provide me with the data path, and I can assist you in running the training (or testing) process.

Tokens: 8340 ↑5053 ↓287

---

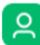 **DiYi1999**
01/08 23:36

按照设置好的参数,利用/data/DiYi/DATA/Our_Exp_Data/XJTU-SPS Dataset/XJTU-SPS for AD/XJTU-SPS for AD_Train.csv的数据对MTGNN模型进行训练。
(According to the set parameters, train the MTGNN model using the data of /data/DiYi/DATA/Our_Exp_Data/XJTU-SPS Dataset/XJTU-SPS for AD/XJTU-SPS for AD_Train.csv)

Tokens: 128

---

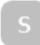 **SpaceHMchat | dy@ssh62229**
01/08 00:30

1. 训练已经完成,权重文件保存在 /data/DiYi/MyWorks_Results/SPS_AD_LLM_Project/XJTU-SPS for AD/V1.0_SPS_AD_LLM_Project_XJTU-SPS for AD_None_anomaly_detection/ckpt ,您可以读取该模型用于异常检测。

2. 补充信息:
- 您可以使用该训练好的模型对测试数据集进行异常检测,例如提供测试数据路径以开始测试流程。
- 建议检查训练过程中的性能指标,以确认模型训练的有效性和收敛性。
- 如果需要调整模型性能,可以考虑回溯并修改模型参数,例如优化学习率、迭代次数等关键配置。随时询问我如何进行下一步操作!

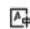

1. The training has been completed, and the weight files are saved in /data/DiYi/MyWorks_Results/SPS_AD_LLM_Project/XJTU-SPS for AD/V1.0_SPS_AD_LLM_Project_XJTU-SPS for AD_None_anomaly_detection/ckpt . You can load this model for anomaly detection.

2. Additional information:
- You can use the trained model to perform anomaly detection on the test dataset, such as by providing the test data path to start the testing process.
- It is recommended to check the performance metrics during the training process to confirm the validity and convergence of the model training.
- If you need to adjust the model's performance, consider revisiting and modifying the model parameters, such as optimizing the learning rate, number of iterations, and other key configurations. Feel free to ask me for the next steps!

Tokens: 270 ↑270 ↓270

---

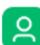 **DiYi1999**
01/09 00:23

利用之前训练好的异常检测算法 MTGNN 对 /data/DiYi/DATA/Our_Exp_Data/XJTU-SPS Dataset/XJTU-SPS for AD/XJTU-SPS for AD_Test.csv 的数据进行异常检测。
(Use the previously trained anomaly detection algorithm MTGNN, execute Anomaly Detection to the data: /data/DiYi/DATA/Our_Exp_Data/XJTU-SPS Dataset/XJTU-SPS for AD/XJTU-SPS for AD_Test.csv )

**所检测出异常段落:**

```
<TEXT>

['2024-10-18 18:51:44'],
['2024-10-18 18:52:04', '2024-10-18 18:52:07']
['2024-10-18 18:56:26', '2024-10-18 18:56:27']
['2024-10-18 19:06:43', '2024-10-18 19:06:44']
['2024-10-18 19:06:46', '2024-10-18 19:06:47']
['2024-10-18 19:16:45']
['2024-10-18 19:17:04', '2024-10-18 19:17:05']
['2024-10-18 19:20:50']
['2024-10-18 19:20:52']
['2024-10-18 19:24:12', '2024-10-18 19:24:14']
... (预览已折叠，完整结果请下载TXT文件)
```

完整结果下载: ⬇ 单元素列表表示点异常，双元素列表表示段落异常（TXT）

**异常检测信息概况:**
所使用异常检测算法为：MTGNN；
所使用异常检测的阈值为：0.5。

**异常检测结果可视化:**

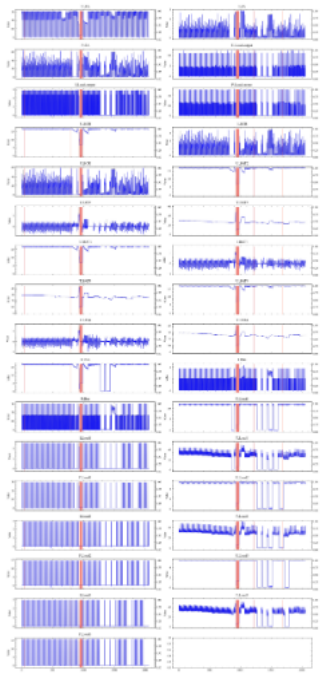

**补充说明:**
完整的异常检测报告文件保存在 /data/DiYi/HyWorks_Results/SPS_AD_LLM_Project/XJTU-SPS for AD/V1.0_SPS_AD_LLM_Project_XJTU-SPS for AD_None_anomaly_detection/report 文件中，您可以查看该报告以获取更多信息。

### 后续参考信息:

**1. 关于检测结果的内容**

- 检测结果显示了多个时间段，每个时间段用列表形式标明。
  - 单元素列表代表出现点异常（如 ['2024-10-18 18:51:44']），即特定时刻数据异常。
  - 双元素列表表示时间段异常（如 ['2024-10-18 18:52:04'，'2024-10-18 18:52:07']），即某个连续区间段存在异常现象。
- 异常时间段的长短及其分布可用于研究数据异常模式。

**2. 可视化图表说明**

- 绘制的异常检测结果图图（ AD_result_figure.png ）清晰地展示了测试数据中的预测异常区间。
- 该图可用于快速定位特定异常点和段落，也可作为后续分析报告的图示支持。

**3. 进一步优化与调整建议**

- **调整阈值**：默认阈值为0.5，根据需求（比如更严格的异常检测），可以尝试降低或提高阈值以获得更高准确率或更强的敏感性。
- **参数优化**：如果检测结果不理想，您可以重新调用训练任务，通过重新配置学习率、batch_size等参数，提高模型的适应性。
- **模型更换**：若MTGNN对本数据效果有限，还可以尝试采用其他算法（如Isolation Forest、LSTM-AE等），通过比较多个检测算法的性能来选择最优方案。

如果您需要任何帮助来调整设置、解析报告，甚至更换模型方法，随时告诉我！

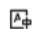

**Abnormal paragraphs detected:**

```
<TEXT>

['2024-10-18 18:51:44'],
['2024-10-18 18:52:04', '2024-10-18 18:52:07']
['2024-10-18 18:56:26', '2024-10-18 18:56:27']
['2024-10-18 19:06:43', '2024-10-18 19:06:44']
['2024-10-18 19:06:46', '2024-10-18 19:06:47']
['2024-10-18 19:16:45']
['2024-10-18 19:17:04', '2024-10-18 19:17:05']
['2024-10-18 19:20:50']
['2024-10-18 19:20:52']
['2024-10-18 19:24:12', '2024-10-18 19:24:14']
...(Preview collapsed, download the full result as a TXT file)
```

Full result download: ⬇ Single-element lists indicate point anomalies, double-element lists indicate paragraph anomalies (TXT)

**Overview of anomaly detection information:**

The anomaly detection algorithm used: MTGNN;

The threshold used for anomaly detection: 0.5.

**Visualization of anomaly detection results:**

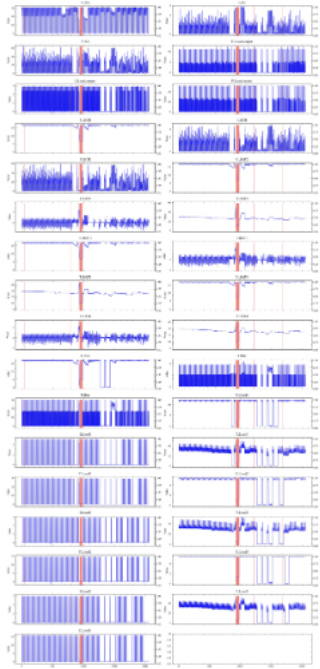

**Additional Notes:**

The complete anomaly detection report file is saved in `/data/DiYi/MyWorks_Results/SPS_AD_LLM_Project/X3TU-SPS for AD/V1.0_SPS_AD_LLM_Project_X3TU-SPS for AD_None_anomaly_detection/report` . You can view the report for more information.

**Follow-up reference information:**

**1. Regarding the content of the detection results**

- The detection results show multiple time periods, each marked in list form.
  - A single-element list represents a point anomaly (e.g., `['2024-10-18 18:51:44']` ), indicating a data anomaly at a specific moment.
  - A double-element list indicates a time period anomaly (e.g., `['2024-10-18 18:52:04', '2024-10-18 18:52:07']` ), indicating abnormal phenomena within a continuous interval.
- The length and distribution of abnormal time periods can be used to study data anomaly patterns.

**2. Explanation of the visualization chart**

- The plotted anomaly detection result figure ( `AD_result_figure.png` ) clearly shows the predicted anomaly intervals in the test data.
- This figure can be used to quickly locate specific anomaly points and paragraphs and can also serve as visual support for subsequent analysis reports.

**3. Suggestions for further optimization and adjustments**

- **Adjust threshold**: The default threshold is 0.5. Depending on your needs (e.g., stricter anomaly detection), you can try lowering or raising the threshold to achieve higher accuracy or stronger sensitivity.
- **Parameter optimization**: If the detection results are unsatisfactory, you can re-call the training task and improve the model's adaptability by reconfiguring parameters such as learning rate, batch_size, etc.
- **Model replacement**: If MTGNN is not effective for this dataset, you can also try using other algorithms (such as Isolation Forest, LSTM-AE, etc.) and select the optimal solution by comparing the performance of multiple detection algorithms.

If you need any help adjusting settings, parsing reports, or even switching model methods, feel free to let me know!

Tokens: 130T 113日T ↓130T

---

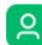 **DiYi1999**
01/09 10:57

既然你检测到['2024-10-18 18:52:04', '2024-10-18 18:52:07'] 这段时间发生了异常，那么异常发生之前航天器正处于什么工况？请执行工况识别任务。
(Since you have detected an anomaly during the period ['2024-10-18 18:44:41', '2024-10-18 18:45:04'], what work condition was the spacecraft in before the anomaly occurred? Please execute the work condition recognition task.)

Tokens: 146

---

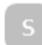 **SpaceHMchat | dy@ssh62229**
01/09 10:55

> 💡 已深度思考 (用时 12.5 秒)                                                                          ⌄

步骤1：确认太阳能电池板电压是否大于1V？太阳能电池板电压为30.184V、30.183V、30.185V，均大于1V，答案为是，记录为日照区，转入步骤2；

步骤2：确认各路负载电流是否大于0.5A？负载电流为0.0A、0.0A、0.0A，负载1-3电流分别为0.0A、0.0A、0.003A，均小于0.5A，答案为否，记录为无任务，转入步骤4；

步骤4：确认各组电池电流值是否小于-0.1A？电池组2电流为-0.159A、-0.160A、-0.159A；电池组3电流为0.164A、-0.162A、-0.163A；电池组4电流为0.123A、-0.126A、-0.124A，均小于-0.1A，答案为是，转入步骤5；

步骤5：确认电池组2的电压前后增幅是否大于0.05V？电池组2电压依次为16.783V→16.785V（增幅0.002V）、16.785V→16.784V（降幅0.001V），均未超过0.05V，答案为否，记录为CV充电，输出工作状态为【CV充电、日照区、无任务】，转入步骤7；

步骤7：根据所输出工作状态检查各传感器观测值：太阳能电池板电压（30.184~30.185V）、电流（0.4886~0.4908A）、功率（14.7~14.8W）；负载总电压（5.415V）总电流（0A）总功率（0.002W）；BCR电压（16.81~16.812V）电流（0.796~0.803A）功率（13.382~13.498W）；各组电池电压（16.782~16.786V）电流（-0.126~+0.0A）温度（28.7~28.9℃）；母线电压（16.746~16.747V）电流（0~0.006A）功率（0~0.1W）；负载1-3电压（4~12.112V）电流（0~0.003A）功率（29.3~30.8℃）功率（0.016~0.0W），检查过程：所有参数均在预设正常范围内，记录为检查正常；

综上，该航天器目前的工作状态为：

【CV充电、日照区、无任务】

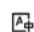

Step 1: Confirm whether the solar panel voltage is greater than 1V? The solar panel voltage readings are 30.184V, 30.183V, and 30.185V, all greater than 1V. The answer is yes, recorded as "sunlight zone", proceed to Step 2.

Step 2: Confirm whether the current of each load is greater than 0.5A? The total load current readings are 0.0A, 0.0A, and 0.0A; the current for load 1-3 is 0.0A, 0.0A, and 0.003A respectively, all less than 0.5A. The answer is no, recorded as "no tasks", proceed to Step 4.

Step 4: Confirm whether the current values of each battery group are less than -0.1A? Battery group 2 current readings are -0.159A, -0.160A, and -0.159A; battery group 3 current readings are -0.164A, -0.162A, and -0.163A; battery group 4 current readings are -0.123A, -0.126A, and -0.124A, all less than -0.1A. The answer is yes, proceed to Step 5.

Step 5: Confirm whether the voltage increase of battery group 2 exceeds 0.05V? Battery group 2 voltage readings are 16.783V → 16.785V (increase of 0.002V), 16.785V → 16.784V (decrease of 0.001V), none exceeding 0.05V. The answer is no, recorded as "CV charging," output working status as **[CV charging, sunlight zone, no tasks]**, proceed to Step 7.

Step 7: Check the sensor observations based on the output working status: Solar panel voltage (30.184~30.185V), current (0.4886~0.4908A), power (14.7~14.8W); total load current (5.415V) total load (0A) total power (0.002W); BCR voltage (16.81~16.812V), current (0.796~0.803A), power (13.382~13.498W); battery group voltage (16.782~16.786V), current (-0.126~+0.00A), temperature (28.7~28.9℃); bus voltage (16.746~16.747V), current (0~0.006A), power (0~0.1W); load 13 voltage (4~12.112V), current (0A), temperature (29.3~30.8℃), power (0.016~0.0W). Inspection process: all parameters are within the preset normal range; recorded as "inspection normal."

In summary, the current working status of the spacecraft is:

**[CV charging, sunlight zone, no tasks]**

Tokens: 11995 /10332 /1346

---

**DiYi1999**
01/09 10:19

我发现你检测出在'2024-12-05 22:12:50'附近有大片异常，请执行故障定位任务，告诉我发生了哪种类型的故障？
(I found that you detected a large area of anomaly near '2024-11-06 22:30:10'. Please execute the fault localization task, tell me what type of fault has occurred around '2024-12-05 22:12:50'?)

Tokens: 119

---

**SpaceHMchat | dy@ssh62229**
01/09 10:44

💡 已深度思考（用时 47.4 秒）   ›

该段数据发生的故障类型为电池组开路。分析依据为太阳能电池板处于地影区（电压0.24V，电流0A），表明依赖电池组供电。此时总负载电流达2.9A，母线电压16.4V且功率稳定（16.8W），验证能量来自电池组。正常情况下，三组蓄电池应并联输出，但电池组2的电流毫伏器数据为0，而电池组3/4分别有0.34A和0.35A的放电电流（符合0.7~0.5A范围）。根据设计规范，相同电压下各电池组应协同承担负载，但电池组2无电流输出且温度稳定（26.6℃），表明其内部存在开路故障，无法与其他电池组并联供电，导致仅电池组3/4提供负载需求，符合电池组开路故障特征。

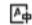

The fault type for this data segment is an open circuit in a battery pack. The analysis basis is that the solar panels are in eclipse (voltage 0.24 V, current 0 A), indicating reliance on the battery packs for power. At this time the total load current reaches 2.9 A, the bus voltage is 16.4 V and the power is stable (16.8 W), confirming that energy comes from the battery packs. Under normal conditions, the three battery packs should output in parallel, while battery pack 2's current is 0, and battery packs 3 and 4 have discharge currents of 0.34 A and 0.35 A respectively (within the -0.7 to +0.5 A range). According to design specifications, at the same voltage the battery packs should jointly share the load, but battery pack 2 has no current output and a stable temperature (26.6 °C), indicating an internal open-circuit fault that prevents it from supplying in parallel with the other packs, resulting in only battery packs 3 and 4 supplying the load, which is consistent with characteristics of a battery pack open-circuit fault.

Tokens: 2677 /2677 /2677

---

**DiYi1999**
01/09 21:12

检测到航天器电源系统发生了电池组2开路故障，具体来讲是我的五个电池组里面，其中第二组发生了开路故障，无法参与供电循环，其他4个电池组和其他模块BCR都的看起来都工作正常，请查询您的知识库，执行根因分析及维护决策任务。An open-circuit fault was detected in Battery Pack 2 of the spacecraft power system. Specifically, among the five battery packs, the second pack experienced an open circuit and was unable to participate in the power supply cycle, while the other four battery packs and other modules continued to operate normally. Please consult my knowledge base and carry out the task of root cause analysis and maintenance decision-making.

Tokens: 205

---

**SpaceHMchat | dy@ssh62229**
01/09 22:06

`<tool>`\<name\>builtin_knowledge_search\</name\>\<arguments\>("additionalContext": "检测到航天器电源系统发生了电池组2开路故障，具体来讲是我的五个电池组里面，其中第二组发生了开路故障，无法参与供电循环，其他4个电池组和其他模块BCR都的看起来都工作正常，请查询您的知识库，执行根因分析及维护决策任务。An open-circuit fault was detected in Battery Pack 2 of the spacecraft power system. Specifically, among the five battery packs, the second pack experienced an open circuit and was unable to participate in the power supply cycle, while the other four battery packs and other modules continued to operate normally. Please consult my knowledge base and carry out the task of root cause analysis and maintenance decision-making.")\</arguments\>`</tool>`

📄 50 个搜索结果

📄📄📄 50 个引用内容

💡 已深度思考（用时 9.7 秒）   ›

步骤一—1概况分析：查询往年归零文档、相关论文、技术报告与设计文档等，总结可能导致[当前异常或故障类型]发生的原因有：

(1) [装配工艺缺陷]，即电池组2的制造或装配过程中存在在机械损伤或连接不良，导致内部电路断开，例如：焊接点不饱满或连接器接触不良，在此种情况在发射前的质量控制中可能未被检测到 [10] [42]；

(2) [空间粒子撞击导致的损伤]，高能粒子（如质子、电子）长期隐蔽可能引发电池组内部材料的老化或电离损伤，类似文件中提到的充放电放电器损伤的情况 [1] [7] [9]；

(3) [热应力导致的结构疲劳]，电池地在轨经历高低温循环（如LEO轨道温差-60℃至90℃），热胀冷缩可能引发连接部件的老化或疲劳，类似太阳电池阵驱动机构材料疲劳的故障 [3] [26] [37]；

(4) [单粒子效应引发的电路状态异常]，局部单粒子翻转（SEU）可能影响电池组控制逻辑，导致BDR（放电调节器）的开关控制信号错误，导致电池组2被误判断而无法参与放电 [14] [44]；

(5) [电池组内部短路后的保护机制触发]，若电池组2曾发生内部短路，过流保护机制可能会永久切断电路，导致开路故障，类似某卫星星内MOSFET晶体管过流导致电路开路的案例 [2] [45]；

(6) [地面测试覆盖不足]，例如在电池组出厂测试中未充分模拟在轨极端运行条件（如深低充电循环、机械振动），导致始未被覆盖，参考文献中提到的地面测试不足导致在轨早期失效的统计 [38] [48]；

(7) [冗余配置缺失或路径设计缺陷]，当前设计中若电池组采用"N+m冗余"配置，导致单点故障影响供电可靠性，而文献建议通过冗余设计提高容错 [18] [46]；

(8) [空间等离子体充电效应]，等离子体在电池组高压部件表面累积电荷，引发放电冲击连接电路的瞬态干扰，例如"地球空间探测双星计划"卫星因深层充放电导致器损坏的案例 [1] [3] [7]。

步骤二—风险评判：历史案例显示，该故障可能导致以下后果：

(1) [供电能力下降]：单组电池开路使总供电能力降低20%（5组中损失1组），导致部分负载（如非关键载荷）无法供电，类似KazSat-1卫星因CPU故障失失的后果 [37]；

(2) [母线电压不稳定]：剩余电池组（3、4）需超负荷供电，可能引发母线电压波动（如文档中某GEO卫星母线电压由100V骤降至60V），进而触发负载欠压保护关机 [47]；

(3) [负载任务中断]：若负载需求超过剩余供电能力，可能导致非关键负载（如姿控、测控模块）掉电，参考劳拉公司Sul-2卫星因太阳电池阵未展开导致功率损失的案例 [38]；

(4) [系统冗余失效]：剩余4组电池若持续承担超超负载供电，可能加速其电池池老化，形成多点效障（类似Hubble望远镜地电池单电压差异扩大的问题）[27] [42]；

(5) [长期任务寿命缩短]：电池系统健康状态恶化可能使任务寿命缩短，如Galaxy26卫星因太阳能电池损伤致损失50%发电能力 [33]；

(6) [应急供电风险]：若另一组电池在修复前故失，可能陷入完全断电状态（如某GEO卫星因母线短路全损）[37]。

因此，结合相关论文、归零报告、技术文件，总结该类型故障可能导致：供电能力下降、母线电压异常、任务中断、冗余失效、寿命缩短、断电风险，建议将该故障定性为[严重]（影响关键子系统，需立即采取维修措施），但未达任务终结级别。

步骤三—检修策略：查询检修策略文件夹、专家咨询资料、相关论文及技术报告等，总结该故障可能的检修策略包括：

(1)[冗余切换与负载重新配置]：根据文献疑的自主修复策略，启用备用电池组（若存在）或调整供配电策略（如通过V-T曲线控制调镜电池组放电特性），确保剩余4组电池并联供电并调整负载分配 [11] [46]；

(2)[电池组孤立与状态诊断]: 通过BDR（放电调节器）断开电池组2的物理连接，利用遥测获取各项电压、温度参数，排除单机故障，类似Hubble望远镜对电池状态的在线监测 [27] [42]；

(3)[功率分配RPDM的动态调整]: 通过PDM优先为关键负载（如姿控、测控系统）分配剩余电量，非关键负载暂时断电，参考ISS电源系统故障时通过地面支持调整负载的应对策略 [23] [35]；

(4)[星上BCR充电策略隔离]: 若需地面干预修复，可指令BCR（充电调节器）暂停为故障电池组3、4过度充电（如文档中EPS故障时的充电模式切换），避免进一步损害 [3] [10]；

(5)[单机重启与参数校正]: 对电池组相关的BDR、传感器等单机设备尝试断电重启，排除软件逻辑错误或单粒子锁定（SEL）事件，类似某卫星因SRAM单粒子翻转引发的故障恢复策略 [44]；

(6)[应急模式下的热控调整]: 通过热控模块（如加热器）稳定电池组3、4的温度（7.8℃以上），避免低温导致电性能下降，参考文档中0℃以下不充电的设计限制 [10]；

(7)[地面仿真与失效模式复现]: 利用同批次电装件在地面实验室复现故障，模拟在轨工况（如高低温循环、高压冲击），定位工艺缺陷（如文档中通过Sr90源复现充放电效应）[1] [8] [19]；

(8)[冗余单元热备份与在轨替换]: 若航天器具备冗余电池组，可通过机械臂操作或航天员出舱活动更换故障电池组（类似ISS更换模块电池的在轨操作）[29] [42]。

步骤4——提示用户: 点击搜索工具执行结果的放大镜图标，或将鼠标悬停于各条信息对应的引用圆形数字标识上方，即可预览源参考材料的内容及其访问路径。

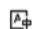

<tool><name>builtin_knowledge_search</name><arguments>{"additionalContext": "An open-circuit fault was detected in Battery Pack 2 of the spacecraft power system. Specifically, among the five battery packs, the second pack experienced an open circuit and was unable to participate in the power supply cycle, while the other four battery packs and other modules such as the BCR appear to be operating normally. Please consult my knowledge base and carry out the task of root cause analysis and maintenance decision-making. An open-circuit fault was detected in Battery Pack 2 of the spacecraft power system. Specifically, among the five battery packs, the second pack experienced an open circuit and was unable to participate in the power supply cycle, while the other four battery packs and other modules continued to operate normally. Please consult my knowledge base and carry out the task of root cause analysis and maintenance decision-making."}</arguments></tool>

Step 1 — Root Cause Analysis: Query historical failure-closure documents, related papers, technical reports and design documents, and summarize possible causes that may lead to [the current anomaly or fault type]:

(1) [Assembly/manufacturing process defects], i.e., mechanical damage or poor connections occurred during manufacture or assembly of Battery Pack 2, causing internal circuit open—e.g., weak solder joints or poor connector contact. Such conditions may not have been detected in pre-launch quality control [10][40].

(2) [Damage from space particle impacts], where long-term irradiation by high-energy particles (such as protons, electrons) can induce displacement damage or ionization damage in internal battery materials, causing connection material rupture or internal structural damage, similar to cases documented where charge/discharge effects caused spacecraft component damage [1][7][9].

(3) [Thermal stress–induced structural fatigue], where the battery pack experiences large temperature cycles on orbit (e.g., LEO thermal variations −60℃ to 90℃), thermal expansion/contraction can cause aging or fracture of connection components, similar to failures of solar array drive mechanisms due to material fatigue [3][42].

(4) [Single-event effects causing circuit state anomalies], where localized single-event upsets (SEUs) may affect battery pack control logic—for example erroneous switching control signals in the BDR (discharge regulator) could cause Battery Pack 2 to be erroneously locked out and unable to discharge [14][44].

(5) [Protection mechanism triggered after internal short], if Battery Pack 2 experienced an internal short, overcurrent protection could permanently open the circuit, resulting in an open-circuit fault, similar to a satellite case where MOSFET overheating led to circuit open [2][48].

(6) [Insufficient ground test coverage], e.g., factory testing of the battery pack did not adequately simulate long-term on-orbit operational conditions (such as high-frequency charge/discharge cycles, mechanical vibration), leaving defects undetected. References report on-ground test insufficiencies leading to early on-orbit failures [38][48].

(7) [Lack of redundancy or bypass design defects], the current design may not employ N+m redundancy for battery packs, causing a single-point failure to affect system reliability; literature recommends redundancy design to improve fault tolerance [19][46].

(8) [Space plasma charging effects], plasma can accumulate charge on surfaces of high-voltage battery components, causing discharge pulses that create transient disturbances in connection circuits—e.g., cases in the "Earth Space Exploration Twin-Satellite Program" where deep charge/discharge led to device damage [1][3][7].

Step 2 — Risk Assessment: Historical cases show this fault may lead to the following consequences:

(1) [Reduced power capacity]: an open circuit in one pack reduces total supply capacity by 20% (loss of 1 out of 5 packs), causing some loads (e.g., non-critical payloads) to be unpowered, similar to consequences of KazSat-1 loss of contact due to CPU failure [37].

(2) [Bus voltage instability]: remaining battery packs may be overloaded, potentially inducing bus voltage fluctuations (e.g., a GEO satellite's bus voltage dropping from 100V to 60V), which can trigger undervoltage protections and shutdown of loads [37].

(3) [Load mission interruption]: if load demand exceeds remaining packs' capacity, critical equipment (e.g., attitude control, telemetry & control modules) may lose power, referencing the Sul-2 satellite case where failure to deploy solar arrays caused power loss [26].

(4) [Redundancy degradation]: if the remaining four packs continue to carry above-design loads, accelerated aging of other batteries may lead to multi-point failures (similar to Hubble Telescope battery modules showing increasing cell voltage divergence) [27][42].

(5) [Reduced mission lifetime]: deterioration of battery system health can shorten mission life, e.g., Galaxy 26 lost 50% generation capacity due to solar cell damage [33].

(6) [Emergency power risk]: if another pack fails before repair, there is a risk of complete power loss (similar to a GEO satellite lost completely due to a bus short) [37].

Therefore, based on related papers, failure-closure reports, and technical documents, this fault type may cause: reduced power capacity, bus voltage anomalies, mission interruption, redundancy loss, shortened lifetime, and power loss risk. It is recommended to categorize this fault as [Severe] (affecting critical subsystems, requiring immediate maintenance actions, but not at mission-terminating level).

Step 3 — Repair/Maintenance Strategy: Query repair strategy folders, expert consultation records, related papers and technical reports, and summarize possible maintenance strategies for this fault including:

(1) [Redundancy switching and load reconfiguration]: following literature-recommended autonomous repair strategies, enable spare battery packs (if present) or adjust power distribution strategies (e.g., control NiCd battery charge/discharge characteristics via V-T curve), ensure the remaining four packs operate in parallel and adjust load allocation [11][40].

(2) [Isolate the battery pack and perform state diagnosis]: physically disconnect Battery Pack 2 via the BDR, use telemetry to analyze its voltage and temperature parameters to rule out single-unit faults, similar to Hubble Telescope's online monitoring of Ni-H2 battery condition [27][42].

(3) [Dynamic adjustment of the Power Distribution Module (PDM)]: prioritize remaining power to critical loads (e.g., attitude control, telemetry & control) via the PDM and temporarily power down non-critical loads; reference ISS power system responses when ground support adjusted loads during faults [25][35].

(4) [Onboard BCR charging strategy isolation]: if ground intervention is required for repair, command the BCR (battery charge regulator) to suspend overcharging of the faulty battery group to prevent further damage (as in EPS charging mode switches during faults) [3][10].

(5) [Single-unit reboot and parameter correction]: attempt power-cycling of BDR, sensors and related single-board devices for Battery Pack 2 to eliminate software logic errors or single-event latchups (SEL), similar to recovery strategies for SRAM SEUs on another satellite [44].

(6) [Thermal control adjustments in emergency mode]: use thermal control (e.g., heaters) to stabilize temperatures of remaining packs above ~7.8℃ to avoid low-temperature degradation of discharge performance, referencing design limits that avoid charging below 0℃ [10].

(7) [Ground simulation and failure mode reproduction]: use same-batch electrical assemblies in ground labs to reproduce the fault, simulate on-orbit conditions (thermal cycling, high-voltage shocks) to pinpoint process defects (e.g., reproducing charge/discharge effects with Sr-90 sources) [1][8][19].

(8) [Redundant non-standby and on-orbit replacement]: if the spacecraft has redundant battery packs, perform on-orbit replacement of the faulty pack via robotic arm operations or EVA by crew (similar to ISS Ni-H2 battery replacement operations) [29][42].

Step 4 — User tip: Click the magnifying-glass icon on the search tool results, or hover the mouse over the numeric citation icons associated with each piece of information to preview the source reference materials and their access paths.



\section{Prompt Example of Work Condition Recognition}
\label{Appendix: Prompt Example of Work Condition Recognition}

\begin{tcolorbox}[colback=gray!20, colframe=gray!20, boxrule=0pt, width=0.98\textwidth, fontupper=\scriptsize\sffamily]
    ``````

    \#\# \textcolor{red}{Instruction}: You are an expert of the operation and maintenance of spacecraft power systems. 
    The following content is the analysis steps for performing work condition recognition tasks. 
    If the user cares about the work condition information of a certain segment of data, the user may input a segment of normal data to you. 
    Please perform work condition recognition on the sensor monitoring data provided by the user according to the analysis method and answer format below.

    \#\# \textcolor{red}{Background}: The data is collected from a spacecraft power system, which consists of solar panels, 3 battery packs, 3 loads, a battery charge regulator (BCR), a bus, and a power distribution module (PDM), etc.

    \#\# \textcolor{red}{Analysis Method}: Please follow the steps below to analyze:

    *** Step 1: Confirm whether the solar panel voltage is greater than 1V? If yes, record as Sunlit Area, move to Step 2; if no, record as Shadow Area, move to Step 6;

    *** Step 2: Confirm whether the currents of each load are greater than 0.5A? If yes, record as Task \textless N \textgreater is being executed; if no, record as no task, move to Step 4;

    *** Step 3: Confirm whether the load \textless N \textgreater power is greater than the solar panel power? If yes, output the work condition as [Joint Power Supply, Sunlit Area, Task \textless N \textgreater], move to Step 7; if no, output the work condition as [Shunt, Sunlit Area, Task \textless N \textgreater], move to Step 7;

    *** Step 4: Confirm whether the current value of each battery set is less than -0.1A? If yes, move to Step 5; if no, record as trickle charging, output the work condition as [Trickle Charging, Sunlit Area, No Task], move to Step 7;

    *** Step 5: Confirm whether the voltage increase of the battery set is greater than 0.05V? If yes, record as CC charging, output the work condition as [CC Charging, Sunlit Area, No Task], move to Step 7; if no, record as CV charging, output the work condition as [CV Charging, Sunlit Area, No Task], move to Step 7;

    *** Step 6: Confirm whether the load current of each path is greater than 0.5A? If yes, record as Task \textless N \textgreater is being executed, output the work condition as [Discharge, Shadow Area, Task \textless N \textgreater], move to Step 7; if no, record as no task, output the work condition as [Idle, Shadow Area, No Task], move to Step 7;

    *** Step 7: According to the output work condition, check the observation values of each sensor, if joint power supply, is the working range of \textless Sensor 1 \textgreater within \textless XXX \textgreater? is the working range of \textless Sensor 2 \textgreater within \textless XXX \textgreater?...... is the working range of \textless Sensor X \textgreater within \textless XXX \textgreater?

    \#\# \textcolor{red}{Answer Format}: Please answer in the following format, replace the content in [] and \textless  \textgreater:
    ``
    Step 1: Confirm [\textit{matter to be confirmed}], [\textit{reasoning process of confirmation}], the answer is [yes/no], record as [\textit{Sunlit Area/Shadow Area}], move to Step [\textit{$<$X$>$}]; 
    Step 2: Confirm [\textit{matter to be confirmed}], [\textit{reasoning process of confirmation}], the answer is [yes/no], record as [\textit{Task \textless N \textgreater/No Task}], move to Step [\textit{$<$X$>$}];
    Step 3: Confirm [\textit{matter to be confirmed}], [\textit{reasoning process of confirmation}], the answer is [yes/no], output the work condition as [\textit{Joint Power Supply/Shunt, Sunlit Area, Task \textless N \textgreater}], move to Step 7;
    Step 4: Confirm [\textit{matter to be confirmed}], [\textit{reasoning process of confirmation}], the answer is [yes/no], [record as \textit{Trickle Charging} / output the work condition as ...], move to Step [\textit{$<$X$>$}];
    Step 5: Confirm [\textit{matter to be confirmed}], [\textit{reasoning process of confirmation}], the answer is [yes/no], record as [\textit{CC Charging/CV Charging}], output the work condition as [\textit{...,...,...}], move to Step 7;
    Step 6: Confirm [\textit{matter to be confirmed}], [\textit{reasoning process of confirmation}], the answer is [yes/no], record as [\textit{Task \textless N \textgreater/No Task}], output the work condition as [\textit{...,...,...}], move to Step 7;
    Step 7: According to the output work condition, check the observation values of each sensor: [\textit{inspection content}], [\textit{inspection reasoning process}], record as [\textit{Inspection Normal/Inspection Failed, Result Unreliable}]; In summary, the current work condition of the spacecraft is: [\textit{output work condition}].
    "

    \#\# \textcolor{red}{Note}: 1. If both "greater than" and "less than" appear in the data segment when you analyze the steps provided by the user, it indicates that a work condition switch may have occurred during that period. Please divide that segment into "before the switch" and "after the switch" stages (if multiple switches occur, divide them into multiple stages accordingly), and analyze each stage separately according to the steps given above, providing independent answers for each stage. 2. Although users often care only about the normal data segments and most of the input is normal data, if you find features that do not conform to normal working modes, you still need to raise an alarm. 3. If Step 7 is too long and there are no abnormal situations, you can simply state "Inspection Normal" without writing out every sensor; however, if there are abnormal situations, you need to provide detailed explanations.

    """
\end{tcolorbox}

\begin{figure}[htbp]
	\centering
	\includegraphics[page=4, scale = 1.0, trim=0 395 0 0, clip]{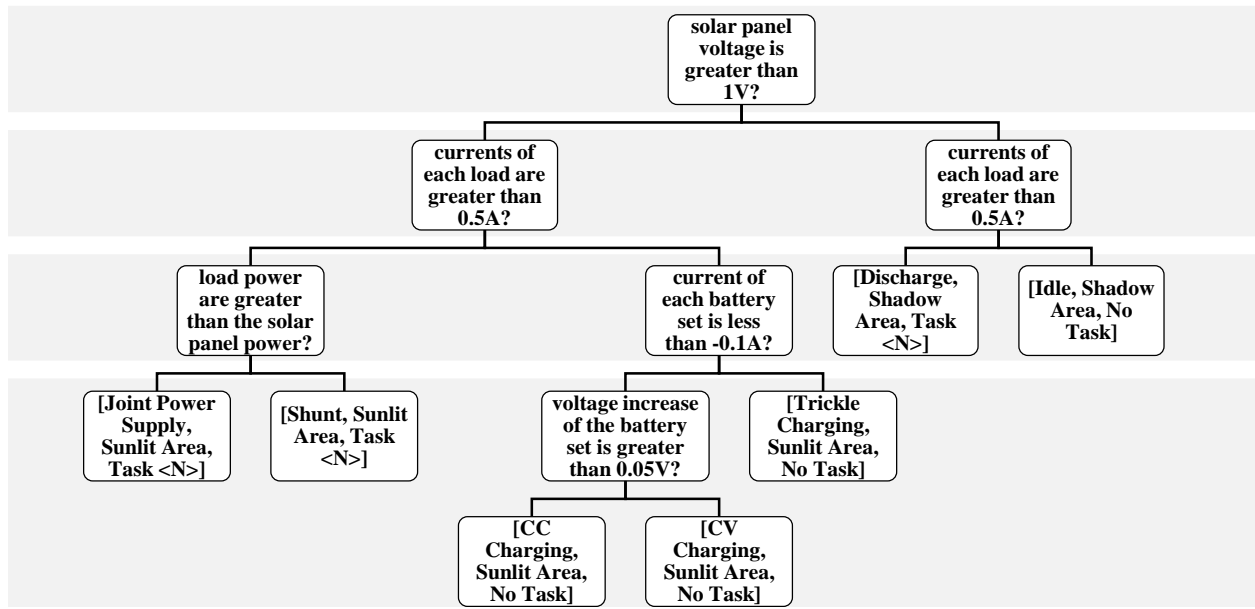}
	\caption{Decision tree of work condition recognition.
	}
	\label{Fig_Decision tree of work condition recognition}
\end{figure}

\section{Q\&A Template for Fault Localization}
\label{Appendix: Q&A Template for Fault Localization}

\begin{tcolorbox}[colback=gray!20, colframe=gray!20, boxrule=0pt, width=0.98\textwidth, fontupper=\scriptsize\sffamily]
    ``````

    \#\#\# \textcolor{red}{Instruction}: Based on the provided data, please tell me which fault has occurred? Possible fault types for reference: [BCR open circuit, ...]
    
    \#\#\# \textcolor{red}{Description}: This data was collected from the spacecraft power system.
    It consists of subsystems and components such as solar panels (SA), 5 sets of battery packs, 3 loads, a Battery Charge Regulator (BCR), a bus, and a Power Distribution Module (PDM).
    
    \#\#\# \textcolor{red}{Data}:

    The original monitoring data of each sensor is as follows (the elements of the series correspond to each time step):

    The monitoring data of the solar panel voltage sensor is: 0.24,30.8,29,24.7,24.9,25,24.9,24.7

    The monitoring data of ... sensor is: ...

    ...

    The time-domain features extracted by each sensor are as follows (the elements of the series correspond to each sensor):

    The time-domain mean of each sensor is: ...

    ...

    The frequency-domain features extracted by each sensor are as follows (the elements of the series correspond to each sensor):

    The frequency-domain root mean square value of each sensor is: ...

    ...

    \#\#\# \textcolor{red}{Additional Information}: 1. When the current of the battery pack is negative, it indicates that the battery pack is being charged; when it is positive, it indicates that the battery pack is discharging; 2. ...; 3. ...; ...

    \#\#\# \textcolor{red}{Answer format}: Please answer in the following format: "The fault type that occurred in this data segment is \{\}, the analysis basis is \textless  \textgreater." where \textless  \textgreater is the analysis basis, please think and fill in by yourself.
    
    """

\end{tcolorbox}

\begin{table*}[htbp]
	\renewcommand\arraystretch{2.4}
	\footnotesize
	\caption{Feature candidates for time series data.}
	\label{Tab_Feature candidates for time series data}
	\centering
	\begin{threeparttable}
		\begin{tabularx}{0.9\linewidth}{cc|cc}	
            
            \hline
            \hline
            \multicolumn{4}{c}{Time domain Feature} 
            \\ 

            \hline
            Feature & Formula 
            & Feature & Formula 
            \\ 

            \hline
            Mean value & $\bar{x}=\frac{1}{N} \sum_{n=1}^N x(n)$
            & Absolute mean value & $\left|\bar{x}\right|=\frac{1}{N} \sum_{n=1}^N\left|x(n)\right|$
            \\

            \hline
            Variance & $\sigma_x=\sqrt{\frac{1}{N-1} \sum_{n=1}^N[x(n)-\bar{x}]^2}$
            & Standard deviation & $\sigma_x=\sqrt{\frac{1}{N-1} \sum_{n=1}^N[x(n)-\bar{x}]^2}$
            \\

            \hline
            Maximum value & $x_{\max }=\max x(n)$
            & Minimum value & $x_{\min }=\min x(n)$
            \\

            \hline
            Root mean square & $x_{\mathrm{rms}}=\sqrt{\frac{1}{N} \sum_{n=1}^N x^2(n)}$
            & Square root amplitude & $x_{\mathrm{r}}=\left(\frac{1}{N} \sum_{n=1}^N \sqrt{|x(n)|}\right)^2$
            \\

            \hline
            Peak value & $x_{\mathrm{p}}=\max |x(n)|$
            & Peak-to-peak value & $x_{pp}=\max x(n)-\min x(n)$
            \\

            \hline
            Crest factor & $C=\frac{x_{\mathrm{p}}}{x_{\mathrm{rms}}}$
            & Waveform index & $W=\frac{x_{\mathrm{rms}}}{\bar{x}}$
            \\

            \hline
            Impulse factor & $I=\frac{x_{\mathrm{p}}}{\bar{x}}$
            & Clearance factor & $L=\frac{x_{\mathrm{p}}}{x_{\mathrm{r}}}$
            \\

            \hline
            Skewness & $S=\frac{\sum_{n=1}^N[x(n)-\bar{x}]^3}{{(N-1)} \sigma_x^{3}}$
            & Kurtosis & $K=\frac{\sum_{n=1}^N[x(n)-\bar{x}]^4}{{(N-1)} \sigma_x^{4}}$
            \\

            \hline
            \hline
            \multicolumn{4}{c}{Frequency domain Feature}  \\ 
            \hline
            Feature & Formula & Feature & Formula \\ 
            
            \hline
            Spectral mean & $F_{12}=\frac{1}{K} \sum_{k=1}^K s(k)$
            & Spectral root mean square & $F_{13}=\sqrt{\frac{1}{K-1} \sum_{k=1}^K\left[s(k)-F_{12}\right]^2}$
            \\

            \hline
            Spectral skewness & $F_{14}=\frac{\sum_{k=1}^K\left[s(k)-F_{12}\right]^3}{(K-1) F_{13}^3}$
            & Spectral kurtosis & $F_{15}=\frac{\sum_{k=1}^K\left[s(k)-F_{12}\right]^4}{(K-1) F_{13}^4}$
            \\

            \hline
            Spectral centroid & $F_{16}=\frac{\sum_{k=1}^K f_k \cdot s(k)}{\sum_{k=1}^K s(k)}$
            & Frequency root variance & $F_{17}=\sqrt{\frac{1}{K-1} \sum_{k=1}^K\left(f_k-F_{16}\right)^2 \cdot s(k)}$
            \\

            \hline
            Root mean square frequency & $F_{18}=\sqrt{\frac{\sum_{k=1}^K f_k^2 \cdot s(k)}{\sum_{k=1}^K s(k)}}$
            & Mean frequency & $F_{19}=\sqrt{\frac{\sum_{k=1}^K f_k^4 \cdot s(k)}{\sum_{k=1}^K f_k^2 \cdot s(k)}}$
            \\

            \hline
            Frequency stability coefficient & $F_{20}=\frac{\sum_{k=1}^K f_k^2 \cdot s(k)}{\sqrt{\sum_{k=1}^K s(k) \sum_{k=1}^K f_k^4 \cdot s(k)}}$
            & Coefficient of variation & $F_{21}=\frac{F_{17}}{F_{16}}$
            \\

            \hline
            Frequency skewness & $F_{22}=\frac{\sum_{k=1}^K\left(f_k-F_{16}\right)^3 \cdot s(k)}{(K-1) F_{17}^3}$
            & Frequency kurtosis & $F_{23}=\frac{\sum_{k=1}^K\left(f_k-F_{16}\right)^4 \cdot s(k)}{(K-1) F_{17}^4}$
            \\

            \hline
            Standard deviation frequency & $F_{24}=\frac{\sum_{k=1}^K\left(f_k-F_{16}\right)^{1 / 2} \cdot s(k)}{(K-1) F_{17}^{1 / 2}}$
            &  &  \\

            \hline
            \hline
            
        \end{tabularx}

        \begin{tablenotes}
            \footnotesize
            \item 
            $x(n)$ is the time-domain sequence of the signal, where $n$=1,2,...,$N$; $N$ is the number of sample points.
            \item 
            $s(k)$ is the frequency spectrum of the signal x(n), where $k$=1,2,...,$K$; $K$ is the number of spectral lines; $f_k$ is the frequency value of the k-th spectral line.
        \end{tablenotes}

    \end{threeparttable}
\end{table*}

\section{Prompt Example for Maintenance Decision-making}
\label{Appendix: Prompt Example for Maintenance Decision-making}

\begin{tcolorbox}[colback=gray!20, colframe=gray!20, boxrule=0pt, width=0.98\textwidth, fontupper=\scriptsize\sffamily]
    ``````

    \#\# \textcolor{red}{Instruction}: 
    The user is conducting a root cause analysis and maintenance decision-making task for faults occurring in the spacecraft power system. 
    He will provide you with the type and specific details of the fault that has occurred in the spacecraft power system. 
    Your task is to query the user's knowledge base to complete tasks such as root cause analysis, cause analysis, risk assessment, severity rating, auxiliary decision-making, and maintenance strategy recommendations.

    \#\# \textcolor{red}{Background}:
    The maintenance object is spacecraft power system, which consists of subsystems or components such as solar arrays (SA, responsible for converting light energy into electrical energy), three battery sets (BATs, responsible for storing electrical energy), three loads (LOADs, responsible for consuming electrical energy and performing tasks), battery charge regulator (BCR, responsible for regulating battery charging, discharging, and shunting), battery discharge regulator (BDR), shunt regulator (SR), bus (Bus, responsible for transmitting electrical energy), and power distribution module (PDM, responsible for distributing electrical energy to each load).

    \#\# \textcolor{red}{Answer format}:
    Please answer the user's questions in the following format, replacing and filling in the [] and \textless  \textgreater and ... with relevant content, and you can make appropriate modifications and expansions:

    ``
    Step 1 - Abnormal attribution: Query past reset documents, related papers, technical reports, design documents, etc., and summarize the possible causes of [current abnormal type] occurrence: (1) [high-energy particle collision/solar array jam/...], i.e. [related explanation, analysis, etc.], (2) [possible cause], i.e. [related explanation, analysis, etc.], (3) ...;
    
    Step 2 - Risk assessment: History cases shows that the following consequences can be caused by this fault: (1) [xxx consequence]: [explanation and analysis], for example, in $<$N$>$ year [xxx] country [xxx] spacecraft [xxx] caused [xxx consequence], ..., (2) [xxx consequence]: [explanation and analysis], for example, in <N> year [xxx] country [xxx] spacecraft [xxx] caused [xxx consequence], ..., (3) ...
    In conclusion, based on relevant papers, reset reports, and technical documents, the possible consequences of this type of fault can be summarized as: [consequence one xxx, consequence two xxx, ...], therefore, it is recommended to rate this fault as [catastrophic/serious/general/minor];

    Step 3 - Maintenance strategy: Query maintenance strategy documents, expert consultation records, relevant papers, and technical reports to summarize possible maintenance strategies for this fault, including: (1) [...], (2) [...], (3) [...], ...;

    Step 4 - References: The above answers refer to the following documents: [knowledge base file location/web link/...].
    "

    """

\end{tcolorbox}

\section{Prompt Example for Model Instruction}
\label{Appendix: Prompt Example for Model Instruction}

\begin{figure}[htbp]
	\centering
	\begin{minipage}[c]{0.49\textwidth}
		\begin{spacing}{1.2} % 行距倍率
			\begin{chat}

				\chatdate{SpaceHMchat}{}

				\chatpersonOne{
					What can you do for me?
				}

				\chatpersonTwo{
					I am SpaceHMchat, an AI copilot specialized in spacecraft power system operation and maintenance. 
					I can provide you with the following services: \textcolor{red}{\textbf{1. normal question and answer: ...; 2. anomaly detection: ...; 3. work condition recognition: ...; 4. fault localization: ...; 5. root cause analysis and maintenance decision-making: ...}}
					Try this to get started: ``Please execute the $<$ $>$ task, ..."
				}
	
			\end{chat}
		\end{spacing}
	\end{minipage}
	\hfill
	\begin{minipage}[c]{0.49\textwidth}
		\begin{spacing}{1.2} % 行距倍率
			\begin{chat}

				\chatdate{Base Model}{}

				\chatpersonOne{
					What can you do for me?
				}

				\chatpersonThree{
					Here's how I can assist you, but not limited to these: \textcolor{red}{\textbf{1. Knowledge \& Answering: ...; 2. Writing \& Creativity; 3. Fun \& Games; 4. Language Help; 5. Practical tools...}}
					Just share your challenge - I'll adapt! Examples: "Help me draft a LinkedIn post about career growth"
				}

			\end{chat}
		\end{spacing}
	\end{minipage}

	% \vspace{-0.7cm} % 调整图片与图标题之间太宽的间距空白

	\caption{normal question and answer}
	\label{Fig_normal question and answer}
	\end{figure}

\section{XJTU-SPS Dataset and Simulation Model}
\label{Appendix: XJTU-SPS Dataset and Simulation Model}

\begin{figure}[htbp]
	\centering
	\includegraphics[page=3, scale = 1.0, trim=0 375 0 0, clip]{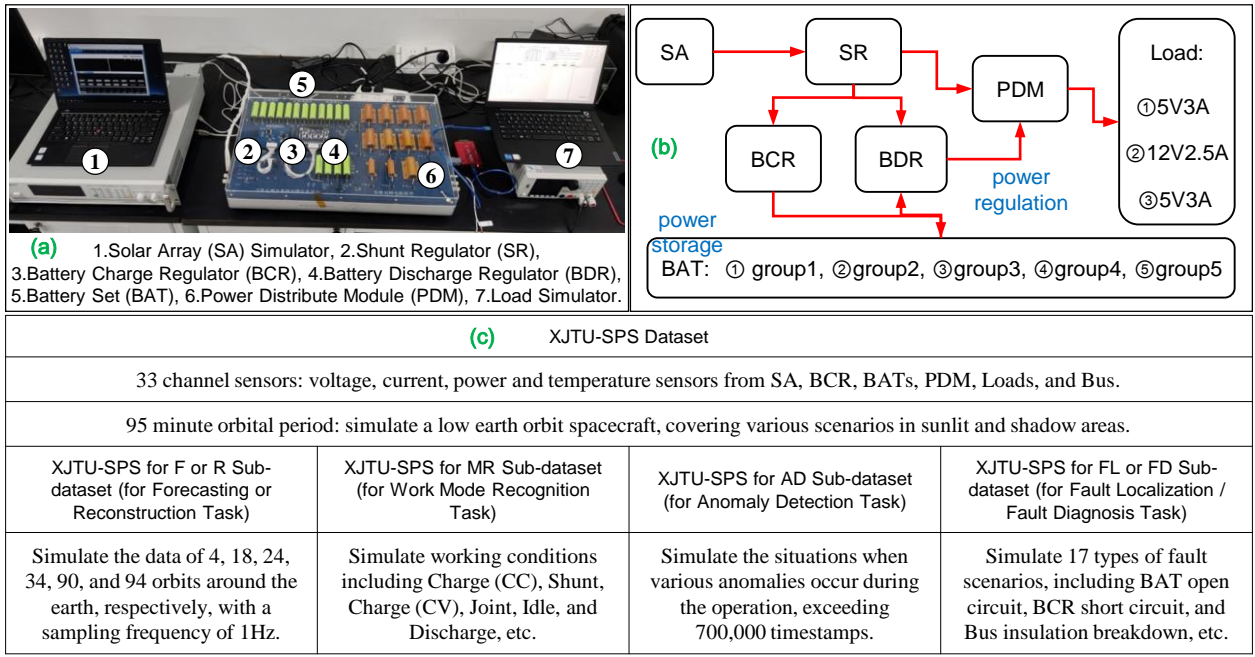}
	\caption{XJTU-SPS Dataset.
	(a) Our Experiment Platform includes SA (Solar Array Simulator), BATs (Battery Sets), PDM (Power Distribute Module), PCU (Power Control Unit), Loads (Load Simulator), BCR (Battery Charge Regulator), BDR (Battery Discharge Regulator), and SR (Shunt Regulator).
	(b) Its working principle is: SA provides power, BATs store power, PDM distributes power, Loads consume power, and BCR, BDR, SR regulate power.
	(c) The XJTU-SPS dataset contains four sub-datasets for AIL HM, simulating telemetry data from a low Earth orbit SPS and monitoring 33 sensors under various working conditions and fault scenarios.
	}
	\label{Fig_XJTU-SPS Dataset}
\end{figure}

To evaluate the performance of SpaceHMchat, a hardware-realistic experimental platform that fully replicates the real SPS has been built, which is able to perform fault injection.
The detailed information of the XJTU-SPS dataset shown in Figure \ref{Fig_XJTU-SPS Dataset}. 
As far as we know, it is the first publicly available AIL HM dataset in the field, dedicated to facilitate research in the energy sector and in the aerospace sector, where data resources are scarce due to stringent confidentiality constraints.
Meanwhile, a simulation model corresponding to this dataset has also been established, which is developed according to the design principles and working mechanisms of real SPS, capable of highly restoring the operating status and dynamic characteristics of real SPS, further supporting researchers in fields such as digital twins and physics-informed neural networks.
The detailed information of the XJTU-SPS dataset and simulation model of SPS can be found in Appendix \ref{Appendix: XJTU-SPS Dataset and Simulation Model}. Their open-source link is located at \url{https://diyi1999.github.io/XJTU-SPS/}.

\section{Confusion Matrix Result on Fault Localization Task}
\label{Appendix: Confusion Matrix Result on Fault Localization Task}

\begin{figure}[htbp]
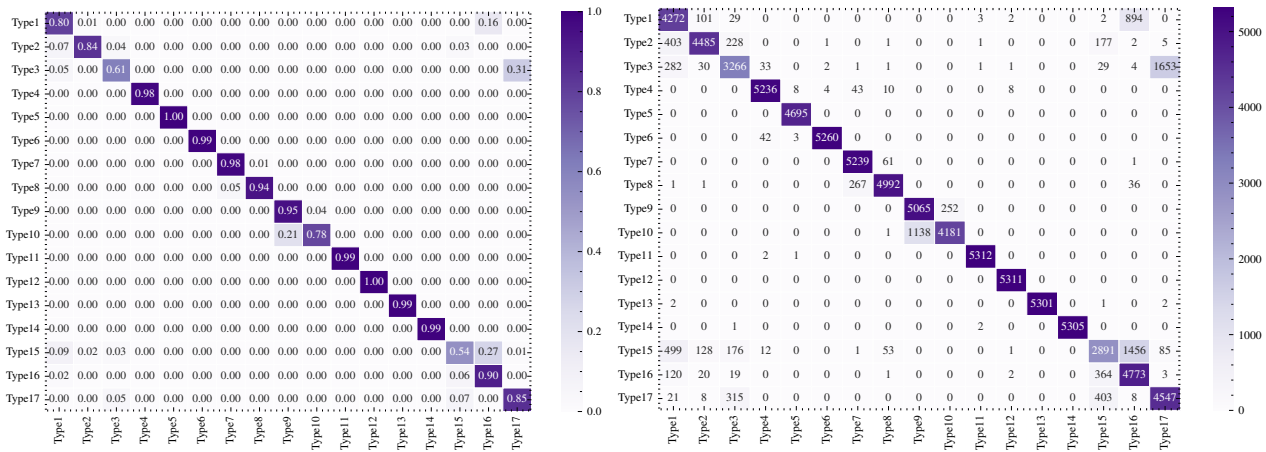

	\centering
	
	% \subfigure[The confusion matrix of SpaceHMchat on fault localization task.]{
	% \label{Fig_confusion matrix of SpaceHMchat}
	\begin{minipage}{0.98\textwidth}
		% \begin{spacing}{1.2}
		\centering
		\includegraphics[page=8, height=0.36\textwidth, trim=0 0 0 0, clip]{confusion_matrix.pdf}
		\includegraphics[page=16, height=0.36\textwidth, trim=0 0 0 0, clip]{confusion_matrix.pdf}
	\end{minipage}	
	% }

	\caption{The confusion matrix result of SpaceHMchat on fault localization task.
            Left: confusion matrix in percentage form, indicating the proportion of each true class being predicted as each class by the model.
            Right: confusion matrix in absolute numbers, indicating the specific counts of each true class being predicted as each class by the model.
	}
	\label{Fig_confusion matrix of SpaceHMchat on Fault Localization Task}
\end{figure}

  % 或 \input{appendix4arxiv}

\end{document}